%% file: main.tex
\documentclass[lettersize,journal]{IEEEtran}
\usepackage{amsmath,amsfonts}
\usepackage{algorithmic}
\usepackage{array}
\usepackage[caption=false,font=normalsize,labelfont=sf,textfont=sf]{subfig}
\usepackage{textcomp}
\usepackage{stfloats}
\usepackage{url}
\usepackage{verbatim}
\usepackage{graphicx}
\usepackage{booktabs}
\usepackage{makecell}
\usepackage{multirow}
\usepackage{pifont}
\usepackage{xcolor}
\usepackage{url}
\usepackage{amsmath}
\usepackage[capitalize,noabbrev]{cleveref}
\usepackage{xcolor}
\usepackage{xspace}
\usepackage{multirow}
\usepackage{colortbl}
\usepackage{wrapfig}
\usepackage{bm}
\usepackage{booktabs}

\newcommand{\eg}{\emph{e.g.}\xspace}
\newcommand{\etal}{\emph{et al.}\xspace}
\newcommand{\ie}{\emph{i.e.}\xspace}

\mathchardef\mhyphen="2D

\usepackage{tikz}
\usetikzlibrary{cd}

\usepackage[T1]{fontenc}%
\usepackage{xhfill}%

\usepackage{pifont}

\usepackage{soul}
\sethlcolor{Gray}

\newcommand{\cut}[1]{}

\def\etal{\textit{et al}. }
\def\ie{\textit{i.e.}}
\def\eg{\textit{e.g.}}

\usepackage{wasysym}

\definecolor{Gray}{gray}{0.9}

\definecolor{Darkgray}{rgb}{0,0,0}
\definecolor{tabgreen}{rgb}{0,0,0}
\definecolor{tabred}{rgb}{0,0,0}
\definecolor{taborange}{rgb}{0,0,0}
\definecolor{tabpurple}{rgb}{0,0,0}
\definecolor{red}{rgb}{1,0,0}
\definecolor{brown}{rgb}{0,0,0}
\definecolor{blue}{rgb}{0,0,0}
\definecolor{magenta}{rgb}{0,0,0}
\definecolor{reviewred}{rgb}{0.8,0.1,0.1}

\newif\ifshowcomments
\showcommentstrue
\ifshowcomments
\definecolor{Gray}{gray}{0.9}
\newcommand{\TODO}[1]{{\color{red}{[TODO: #1]}}}

\else
\newcommand{\TODO}[1]{}
\newcommand{\revised}[1]{}
\fi

\def\BibTeX{{\rm B\kern-.05em{\sc i\kern-.025em b}\kern-.08em
    T\kern-.1667em\lower.7ex\hbox{E}\kern-.125emX}}
\usepackage{balance}

\begin{document}
\title{Vision-Assisted Foundation Model for Solving Multi-Task Vehicle Routing Problems}
%\author{Shuangchun Gui, Zhiguang Cao, ~\IEEEmembership{Senior Member, IEEE}, Wen Song,~\IEEEmembership{Senior Member, IEEE}, and Yew-Soon Ong,~\IEEEmembership{Fellow, IEEE} 

\author{Shuangchun Gui, Zhiguang Cao, Wen Song, and Yew-Soon Ong 

%\thanks{This research is supported by the National Research Foundation, Singapore, under its AI Singapore Programme (AISG Award No: AISG3-RP-2022-031). This research is also partly supported by the MTI under its AI Centre of Excellence for Manufacturing (AIMfg) (Award W25MCMF014), partly by the A*STAR-NTU Research Joint Lab for Smart and Sustainable Advanced Manufacturing, and partly by the National Natural Science Foundation of China under Grant 62473233. \emph{(Corresponding author: Wen Song.)}
%}
\thanks{Shuangchun Gui and Zhiguang Cao are with the School of Computing and Information Systems, Singapore Management University, Singapore (Emails: gshuangchun@outlook.com, zgcao@smu.edu.sg).}
\thanks{Wen Song is with the Institute of Marine Science and Technology, Shandong University, China (Email: wensong@email.sdu.edu.cn).}
\thanks{Yew-Soon Ong is with the College of Computing and Data Science, Nanyang Technological University, Singapore, and the Centre for Frontier AI Research, Institute
of High Performance Computing, Agency for Science, Technology and Research, Singapore (Email: asysong@ntu.edu.sg).
}
}

\markboth{Journal of \LaTeX\ Class Files,~Vol.~18, No.~9, June~2026}%
{How to Use the IEEEtran \LaTeX \ Templates}

\maketitle

\input{0-abs}

\input{1-intro}

\input{2-relat}

\input{3-metd}
\input{4-exp}

\input{5-dis}

\input{6-con}

\bibliographystyle{IEEEtran.bst}
\bibliography{ref.bib}

\end{document}

%% file: 0-abs.tex
\begin{abstract}
Multi-task vehicle routing problems (VRPs) play a critical role in enhancing efficiency across various industries and service sectors. These problems consist of multiple variants that optimize routing costs while meeting diverse customer constraints.
Existing multi-task VRP solvers solely utilize a graph-based modality, limiting their ability to address variants with multiple constraints. 
As a format to represent complex semantics, vision modality shows great potential for encoding diverse VRP constraints. This motivates us to learn patch-level semantics from the vision images, and then integrate them into a graph-based model to solve various VRP variants simultaneously. However, directly applying this approach to multi-task VRPs presents three challenges: 1) existing VRP images lack constraint representations, which are essential for multi-task VRPs, 2) the fixed receptive field of individual patches cannot effectively accommodate varying requirements across tasks, and 3) imbalanced pixel distribution among constraints may cause the model to overlook constraints with fewer pixels.
In this paper, we propose a vision-assisted foundation model (VaFM) to address these challenges. 
In the vision modality, input images tailored to all constraints are encoded by a convolutional neural network (CNN). The obtained patch embeddings are fused with graph-based nodes to generate solutions, with an auxiliary task designed to address the pixel-imbalanced issue. 
Specifically, we design a hybrid cross-attention fusion module to enable adaptive receptive fields for different tasks. It incorporates feature maps from shallow layers to focus on local details and uses cross-patch attention to capture global information.
Moreover, we design a constraint-aware auxiliary task and utilize binary cross-entropy (BCE) loss to ensure balanced learning of all constraints.
The performance of VaFM is evaluated across 16 different VRP variants. The experimental results demonstrate the superiority of VaFM over state-of-the-art (SOTA) methods, especially for variants with complex constraints.
\end{abstract}

% Note that keywords are not normally used for peerreview papers.
\begin{IEEEkeywords}
Multi-task learning, multi-modality model, deep reinforcement learning, vehicle routing problems.
\end{IEEEkeywords}

%% file: 1-intro.tex
% \IEEEraisesectionheading{\section{Introduction}\label{sec:introduction}}
% 
\section{Introduction}
{Vehicle} Routing Problems (VRPs) focus on minimizing routing costs by efficiently managing vehicles to meet customer demands. It plays a critical role in improving efficiency in logistics, transportation, retail distribution, waste collection, and manufacturing~\cite{toth2014vehicle}. However, VRPs are NP-hard combinatorial optimization problems in general, and efficiently solving large-scale instances has been a long-standing challenge in both research and practice. Recently, to accelerate VRP solving, neural VRP methods have received much attention~\cite{joshi2019efficient,kool2018attention,kwon2020pomo,wu2021learning,ma2024learning,sun2023difusco}. Leveraging the representation learning power of modern deep neural networks, these methods extract useful patterns and policies from historical VRP instances through offline learning, so as to accelerate online instance solving. They typically consider VRP instances as graphs, where customer locations are represented as various nodes. VRP models are often designed based on graph neural networks (GNNs)~\cite{joshi2019efficient,sun2023difusco} and transformers~\cite{kool2018attention,kwon2020pomo,luo2023neural}, both of which are capable of learning node-specific features and effectively capturing complex relationships within graph structures.

\begin{figure}[t]%
\centering
    \includegraphics[width=1.0 \columnwidth]{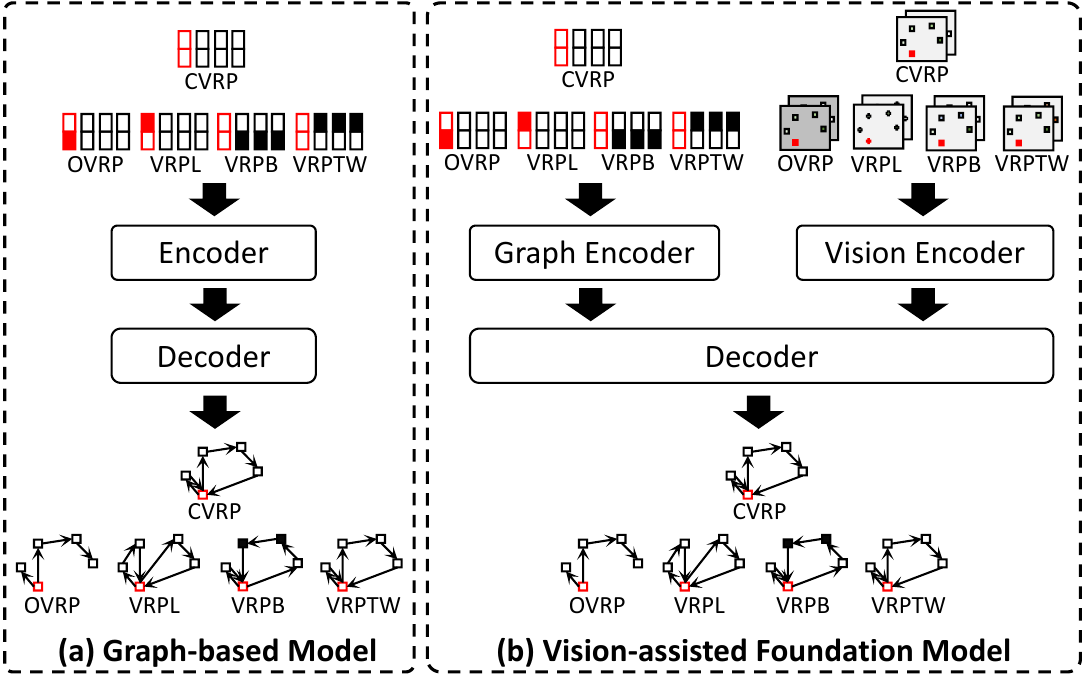}%
    \caption{ (a) Conventional multi-task VRP solvers utilize graph-based attributes as input to an encoder-decoder architecture, which generates next-node selection probabilities. For the four constraints O, L, B, and TW, the input attributes include coordinates, open route flags, distance limits, demands, and time window values. (b) VaFM combines graph and vision modalities to collaboratively generate final solutions. \textcolor{red}{Red} denotes the depot nodes, while Black represents customer nodes.
    }
    \label{fig:intro-model}%
\end{figure}

Despite the promising results, mainstream approaches in neural VRP solving suffer from a major limitation that the solver is designed and trained for a specific VRP variant. It is well known that, in reality, VRP could involve numerous variants with different constraints~\cite{vidal2020concise}. Specifically, given a set of constraints, the number of possible VRP variants may grow exponentially. For example, if we consider four commonly studied constraints, {\textit{backhaul demands} (B) for pickups and deliveries, \textit{open routes} (O) that do not require a return to the depot, \textit{duration limits} (L) on vehicle travel time, and \textit{time windows} (TW) for specific service periods~\cite{braekers2016vehicle,vidal2020concise},} there will be $2^4=16$ VRP variants. If we train a specific model for each variant (or task) separately, as in mainstream approaches, the computational and resource demand will be prohibitive, thereby limiting its feasibility in practical applications~\cite{li2022overview}.
A more reasonable approach is to train a single unified model to capture shared knowledge across different tasks. {For example, Liu~\etal~\cite{liu2024multi} explore this direction by training a model on multiple tasks with single constraints and applying the model to all possible variants.} Moreover, a recent study~\cite{berto2024routefinder} focuses on developing a foundation model for all task variants by using a mixed batch training strategy. These \emph{multi-task} solvers follow the classic graph-based paradigm and employ attributes of the aforementioned constraints (B, O, L, and TW) to formulate node information. 
\color{black}
Specifically, these solvers sequentially select nodes and link them into complete routes under a reinforcement learning (RL) framework. As shown in Fig.~\ref{fig:intro-model}~(a), multiple attributes are fed to an encoder-decoder architecture to generate probabilities for selecting the next node. The resulting action (node) then updates the environment (\eg, remaining capacity and current time) for the subsequent decision step.
While achieving desirable performance, the graph-based solver relies on the model to learn local spatial relations, such as surrounding node density and relative demand distribution. As the task becomes more complex (\eg, VRPs involving both demand objectives and time window restrictions), these relations are difficult for the model to reliably learn without explicit spatial cues.
\color{black}

As a format for intuitively representing complex semantics, vision images demonstrate significant potential in object detection and segmentation~\cite{liu2023image, nguyen2025survey, ye2021continuation, liu2023contrastive, du2022swinpa, wang2023one}.
\color{black}
This motivates us to integrate visual representations into the graph-based VRP solver to enhance its performance.
\color{black}
\textbf{The key idea is to learn patch-level semantics from the vision images, such as spatial demand distributions or service time patterns, and then adopt them to enrich constraint representation for graph-based node embeddings.}
In neural combinatorial optimization, several studies~\cite{ling2020solving,ling2023deep,graikos2022diffusion,samizadeh2023vn} develop vision-based models for solving a single-task VRP, \eg, the traveling salesman problem (TSP). By leveraging vision architectures, such as convolutional neural network (CNN)~\cite{ling2020solving,ling2023deep} and diffusion models~\cite{graikos2022diffusion}, these neural methods learn representations from input images that contain customer location information. 
However, directly applying this approach to a multi-task neural VRP solver presents three challenges: 1) existing VRP images lack crucial attributes, such as task-specific demands and time windows, which are essential for multi-task VRPs, 2) the fixed receptive field of individual patches cannot effectively accommodate varying requirements across tasks, and 3) imbalanced pixel distribution among constraints may cause the model to overlook constraints with fewer pixels.

\begin{figure}[t]%
\centering
    \includegraphics[width=1.0 \columnwidth]{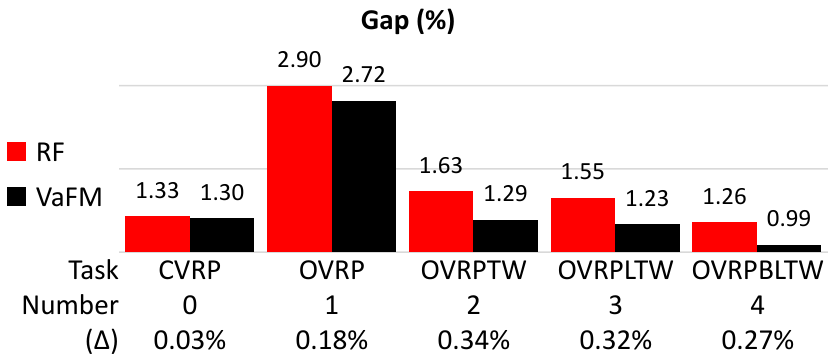}%
    \caption{The performance (gap to solutions found by the SOTA heuristic HGS) of the SOTA method (RF) and the proposed VaFM across different tasks. "RF" and "VaFM" correspond to RF-MoE-L and VaFM-MoE-L in Table~\ref{tab: sota}, respectively. "Number" indicates the number of constraints in each task, while "($\Delta$)" represents the improvement of VaFM over RF.
    }
    \label{fig:intro-gap}%
\end{figure}

To this end, we develop a \underline{\textbf{V}}ision-\underline{\textbf{a}}ssisted \underline{\textbf{F}}oundation \underline{\textbf{M}}odel (VaFM) to address these challenges. As illustrated in Fig.~\ref{fig:intro-model} (b), VaFM combines graph and vision modalities to collaboratively generate final solutions.
In the vision modality, two images are designed to address \textit{challenge 1)}. Both images consist of varying node color intensities, with one representing demand and the other capturing time window information. Additionally, distinct background colors and node shapes are used in both images to represent O and L.
To process the two modality inputs, we employ transformer-based and CNN-based models as the graph and vision encoders, respectively. The obtained graph nodes and vision patches are fused to produce the final solution, with an auxiliary task introduced to mitigate the pixel imbalance issue.
For \textit{challenge 2)}, we propose a hybrid cross-attention fusion module to enable adaptive receptive fields for different tasks. This module collects feature maps from various vision encoder layers and then integrates them into node embeddings using a cross-patch attention mechanism. On the one hand, embeddings from shallow layers have a smaller patch size, which can provide fine-grained context information. On the other hand, cross-patch attention calculates the similarities between node and patch embeddings within an instance, adaptively enlarging the receptive field by aggregating relevant patches for the current node.
Moreover, to address \textit{challenge 3)}, we design an auxiliary task based on constraint recognition. This is achieved by incorporating a classifier after the vision encoder and applying the binary cross-entropy (BCE) loss to guide model training. By encouraging the model to recognize the presence of each constraint, the BCE loss ensures balanced learning across all constraints, regardless of their pixel distribution.

We conduct experiments with various combinations of the four constraints B, O, L, and TW, resulting in 16 VRP variants. Experimental results show that VaFM outperforms SOTA methods, particularly for variants with complex constraints. As illustrated in Fig.~\ref{fig:intro-gap}, VaFM achieves greater improvements on tasks with multiple constraints, such as OVRPTW, OVRPLTW, and OVRPBLTW, compared to simpler tasks like CVRP and OVRP.
Our contributions are summarized as follows:
\begin{itemize}
    \item To the best of our knowledge, our work is the first to develop a vision-assisted foundation model for multi-task VRPs, which combines graph and vision modalities to understand complex attributes.
    \item We propose multi-task VRP images, a hybrid cross-attention fusion module, and a constraint-aware auxiliary task to address the challenges in applying the vision modality to the graph-based VRP solver.
    \item We conduct extensive experiments, and the results show that VaFM surpasses SOTA methods across most variants, especially for those with complex constraints.
\end{itemize}

The remainder of this paper is organized as follows. Section 2 reviews neural methods for VRPs and examines studies on multi-modality learning. Section 3 introduces the conventional pipeline of neural methods for multi-task VRPs. Section 4 describes the architecture of the proposed vision-assisted foundation model. Section 5 presents experimental results and analysis. Section 6 provides additional insights and discussions. Section 7 concludes the paper.

%% file: 2-relat.tex
\section{Related Works}

{We begin by reviewing the graph-based neural VRP solvers, which is followed by an examination of vision-based neural methods for single-task VRPs and the Hamiltonian cycle problem. We further explore the application of multi-modality learning and analyze its mainstream techniques.}
\subsection{Graph-based Neural Solver}
% 加一下简称的全程 斜体全称
% 经典的VRP问题, construction, improvement
We focus on graph-based VRP neural solvers that construct solutions in an autoregressive manner. They typically model VRP instances as graphs, with customer locations represented as nodes. The employed models can be categorized into graph neural networks (GNNs)~\cite{joshi2019efficient} and transformers~\cite{kool2018attention,kwon2020pomo,luo2023neural}.
Joshi~\etal~\cite{joshi2019efficient} exploit graph convolutional networks to model structural relationships between TSP customers. For transformer-based approaches, Nazari~\etal~\cite{nazari2018reinforcement} first adapt pointer networks~\cite{vinyals2015pointer} to VRPs. The attention model (AM)\cite{kool2018attention} employs a transformer with reinforcement learning\cite{williams1992simple} and a rollout baseline to solve VRPs. Kwon~\etal~\cite{kwon2020pomo} further enhance performance with policy optimization using multiple optima (POMO), which leverages diverse rollouts and data augmentations. \textcolor{black}{Recently, DAR~\cite{wang2025distance} proposes distance-aware attention reshaping to improve the size generalization of neural solvers.}
However, these methods are trained and evaluated on the same task, limiting their generalization capability across different task variants.
Multi-task VRPs extend conventional VRP by considering diverse tasks within a single framework, where different combinations of constraints define distinct task variants. These constraints include \textit{backhaul demands} (B) for pickups and deliveries~\cite{zong2022mapdp,kong2024efficient}, \textit{open routes} (O) for routes without a return to the depot~\cite{tyasnurita2024constructing,bezerra2023general}, \textit{duration limits} (L) for travel time restrictions~\cite{oliveira2024hybrid}, and \textit{time windows} (TW) for specific service periods~\cite{zhang2022learning,lin2021deep}. 
\color{black}
Some works train a unified model to capture the shared knowledge~\cite{lin2024cross,liu2024multi,zhou2024mvmoe,berto2024routefinder,berto2025routefinder,drakulic2025goal}. 
\color{black}
For example, Lin~\etal~\cite{lin2024cross} demonstrate a pre-trained TSP model that can be fine-tuned for other VRP tasks. To expand task coverage, recent works~\cite{liu2024multi,zhou2024mvmoe} extend CVRP by integrating B, L, O, and TW. These models are trained on single-attribute VRPs and generalized to various attribute combinations. A recent study~\cite{berto2024routefinder} introduces a foundation model for all task variants, fine-tuned on an unseen attribute (\ie, mixed backhaul). \textcolor{black}{A subsequent extension~\cite{berto2025routefinder} further develops the model architecture by adopting a modern transformer-based encoder.} However, these multi-task VRP solvers rely solely on graph-based inputs, restricting their ability to capture complex constraint interactions.

\subsection{Vision-based Neural Solver}
Motivated by the success of computer vision, some studies investigate the application of the vision modality in combinatorial optimization. Most of them~\cite{ling2020solving,ling2023deep,graikos2022diffusion} employ CNNs and diffusion models to transform a fully connected graph into solution images, aiming to maximize the similarity between the output and the optimal solution image.
For example, Ling~\etal~\cite{ling2020solving,ling2023deep} use convolutional layers to generate solutions, with the latter addressing scale limitations by incorporating multiple repeated sub-problems. Graikos~\etal~\cite{graikos2022diffusion} generate a grayscale image from an adjacency matrix and encode the image using diffusion models.
To intuitively interpret graph-based data, Samizadeh~\etal~\cite{samizadeh2023vn} explore vision-based approaches for solving the Hamiltonian cycle problem. They convert nodes into images with elliptical, spiral, or random distributions, then use a vision backbone followed by a classifier to determine if the input image contains a Hamiltonian cycle.
On the one hand, none of the above works has yet developed a vision-based solver for multi-task VRPs. On the other hand, existing VRP images capture only location information, neglecting the various attributes crucial for constraint understanding. To fill this gap, this paper proposes a vision-based foundation model for solving multiple VRP variants simultaneously. Unlike the single-modality methods, we integrate vision images into a graph-based model to enhance performance.

\subsection{Multi-modality Learning}
% LLM的效果不好
Humans perceive the world by interpreting high-dimensional inputs from diverse modalities, such as text, audio, and visual data~\cite{smith2005development}. Inspired by this, numerous studies explore the application of multi-modality learning (MML) to tasks like speech recognition~\cite{harwath2016unsupervised}, image segmentation~\cite{ye2019cross}, and numerous smart city scenarios~\cite{NEURIPS2024_15cc8e4a}. In the VRP field, efforts have been made to incorporate the vision modality into large language models (LLMs) to enhance their ability to solve routing problems~\cite{huang2021makes}. {However, their performance remains inferior to that of graph-based solvers. In contrast, this paper addresses multi-task VRPs by integrating the vision modality with graph-based models, thereby significantly enhancing the neural solver's capacity.}
As a core aspect of MML, multi-modal fusion integrates information from diverse modalities. Common approaches combine representations from individual models using operations like concatenation~\cite{liu2022concise}, element-wise averaging~\cite{shutova2016black}, and linear combination~\cite{anastasopoulos2019neural}. These operations can occur at different stages of the learning process: early fusion~\cite{perez2019mfas} integrates low-level features, late fusion~\cite{morvant2014majority} combines high-level features, and hybrid fusion~\cite{anastasopoulos2019neural} benefits from both early and late fusion. Due to the substantial heterogeneity between modalities, some tasks require feature alignment to match corresponding elements, such as linking textual descriptions to object regions in images. {Recently, attention mechanisms have become a popular approach for aligning multi-modal features~\cite{cui2023deep}. In this paper, to accommodate adaptive receptive fields for various tasks, vision and graph modalities are fused in a hybrid manner, with node-image features aligned through cross-patch attention.}

%% file: 3-metd.tex
\section{Preliminaries}
\begin{figure}[t]%
\centering
    \includegraphics[width=1.0 \columnwidth]{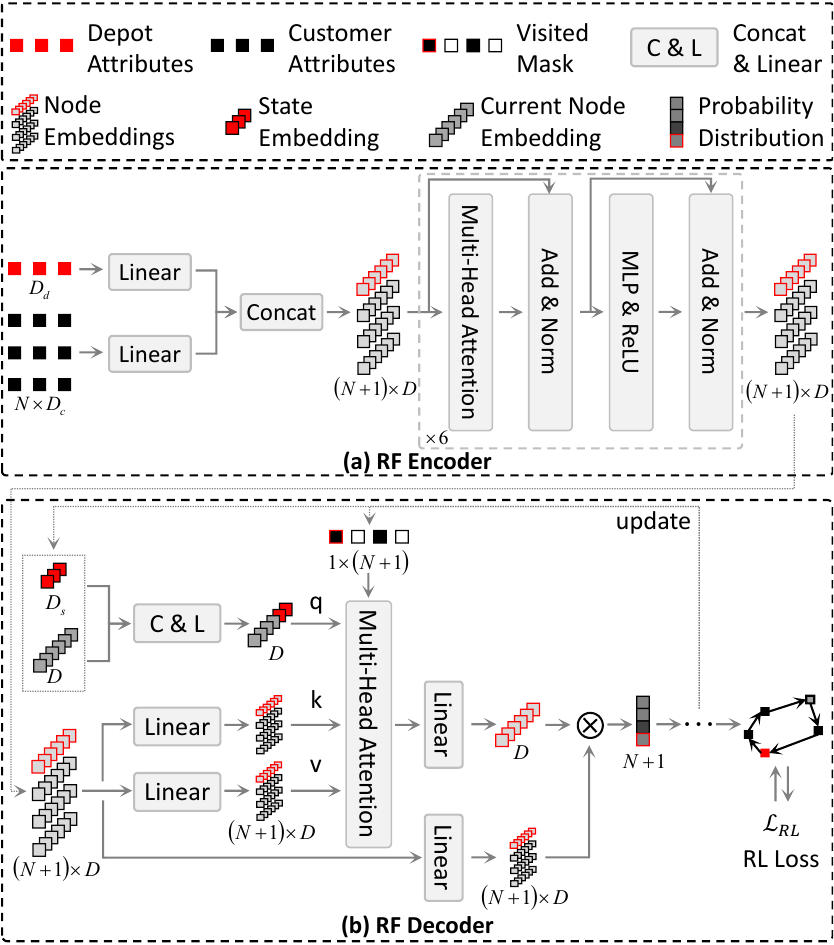}%
    \caption{ (a) Attributes of the depot and customers are passed through linear layers and concatenated to form the initial node embeddings. These embeddings are then encoded by a 6-layer transformer-based model to obtain graph-based node representations.
    \color{black}{(b) The decoder constructs the solution in an auto-regressive manner. At each step, the query is derived from the current node and state embeddings, while the keys and values are obtained from the graph-based node embeddings. Both the encoder and decoder are jointly trained through an RL loss $\mathcal{L}_{RL}$ defined in Eq.~\ref{eq: rl}}.
    }
    \label{fig: RF}%
\end{figure}
\color{black}
\subsection{Problem Definition}

We begin by introducing a typical VRP, \ie, the capacitated vehicle routing problem (CVRP). CVRP aims to determine optimal sub-routes that minimize the total travel cost while satisfying customer demands. In each sub-route, the vehicle must load goods at the depot and deliver them to customers, while ensuring the following three constraints: 1) the sub-route starts and ends at the depot, 2) each customer is visited exactly once, and 3) the total demand of the sub-route does not exceed the vehicle's capacity.
To formulate this problem, we define a graph with different nodes to represent the depot and $N$ customers, denoted as  $\mathbf{V}=\{v_0, v_1, \dots, v_N\}$.
Each customer node has a demand value, and each sub-route is limited by the capacity $Q$. CVRP can be extended to various VRP variants by incorporating additional constraints. Below, we describe each constraint in detail.
Following~\cite{berto2024routefinder}, this paper examines two types of constraints: instance-based \textit{open route} (O) and \textit{duration limit} (L), as well as customer-based \textit{backhaul demand} (B) and \textit{time window} (TW).
Specifically, \textbf{O} is represented by a binary flag $o \in \{0, 1\}$, and in OVRP (\ie, $o=1$), vehicles do not return to the depot $v_0$ after completing a sub-route. Meanwhile, \textbf{L} (\eg, VRPL) imposes a travel length limit $l$ on sub-routes to ensure a balanced workload.
In tasks involving \textbf{B} (\eg, VRPB), customer nodes are classified as linehaul or backhaul. The demands are represented by two sets: $\{\delta_1^l, \dots, \delta_N^l\}$ for linehaul and $\{\delta_1^b, \dots, \delta_N^b \}$ for backhaul, with each demand satisfying $\delta_i^l, \delta_i^b \in [0, Q]$. Linehaul refers to transporting goods from the depot to customers, while backhaul involves returning goods from customers to the depot. Each sub-route must serve all linehaul customers before handling backhauls. 
In tasks with \textbf{TW} (\eg, VRPTW), {each customer $v_i$ has the early and late time windows, as well as a service time $\{t_i^e, t_i^l, t_i^s\}$.} Vehicles must arrive at $v_i$ before $t_i^l$, with early arrivals waiting until $t_i^e$.

\subsection{Conventional Pipeline for Solving Multi-Task VRPs}
We describe the conventional pipeline of neural methods for solving multi-task VRPs, \ie, RouteFinder (RF)~\cite{berto2024routefinder}. As shown in Fig.~\ref{fig: RF}, RF undertakes three steps: 1) encoding input attributes into node embeddings, 2) decoding these embeddings into action probabilities, 3) optimizing a reinforcement learning (RL) objective to minimize routing cost.

\subsubsection{Encoder}
\label{sec: RF-enc}
As shown in Fig.~\ref{fig: RF} (a), the input attributes are denoted as $\{ \mathbf{h}_0, \mathbf{h}_1,\dots, \mathbf{h}_N \}$. The depot attribute $\mathbf{h}_0 = \{c_0^x, c_0^y, o, l\} \in \mathbb{R}^{D_d}$ includes the depot coordinates, open route flag, and duration limit, with a feature dimension of $D_d$. Customer features $\{\mathbf{h}_1, \mathbf{h}_2, \dots, \mathbf{h}_N\} \in \mathbb{R}^{N \times D_c}$ are $D_c$-dimensional, where the $i$-th node $\mathbf{h}_i = \{[c_i^x, c_i^y], [\delta_i^l, \delta_i^b], [t_i^e, t_i^l, t_i^s]\}$ specifies the coordinates, demands, and time windows.
\color{black}
The depot and customer attributes are projected to dimension $D$ through two linear layers $\mathcal{H}(\cdot)$, forming the initial node embedding:
\begin{equation}
    \mathbf{I}= \texttt{Concat}(\mathcal{H}(\mathbf{h}_0),\mathcal{H}(\{\mathbf{h}_1, \mathbf{h}_2,\dots, \mathbf{h}_N\})),
\end{equation}
where $\texttt{Concat}(\cdot)$ denotes the concatenation operation, \ie, a $(1 \times D)$ vector is concatenated with an $(N \times D)$ matrix along the row dimension, yielding $\mathbf{I} \in \mathbb{R}^{(N+1) \times D}$.
\color{black}
{The embedding $\mathbf{I}$ is fed into a 6-layer transformer-based graph encoder $\mathcal{E}_g(\cdot)$, and the node embeddings are derived from the last layer, denoted as $\mathbf{H} = \mathcal{E}_g(\mathbf{I})$, where $\mathbf{H} \in \mathbb{R}^{(N+1) \times D}$.}

\subsubsection{Decoder}
\label{sec: RF-dec}
As shown in Fig.~\ref{fig: RF} (b), the decoding process uses a multi-head attention mechanism to compute the action probability distribution and construct solutions autoregressively across $N$ solution trajectories. For clarity, trajectory indexes are omitted in this subsection.

At step $j$, the current node index is denoted as $\pi_{j}$, and its corresponding node embedding is $\mathbf{H}_{\pi_{j}} \in \mathbb{R}^{D}$. The state embedding, denoted as $\mathbf{S}_{j}=\{ c_{j}, d_{j}, t_{j}\} \in \mathbb{R}^{D_s}$, consists of the remaining capacity $c_{j}$, remaining distance $d_{j}$ (for constraint \textbf{L}), and current time $t_{j}$. 
\color{black}
The current node (1$\times$ D) and state embeddings ($1\times$ $D_s$) are concatenated to $1\times (D+D_s)$, thereby forming the query $\mathbf{q}_{j}= \mathcal{H}(\texttt{Concat}(\mathbf{H}_{\pi_{j}}, \mathbf{S}_{j}))$. $\mathcal{H}(\cdot)$ is a linear layer that maps the features from $D_s + D$ to $D$, resulting in $\mathbf{q}_{j} \in \mathbb{R}^{D}$.
\color{black}
In the multi-head attention mechanism, the key and value embeddings are learned from the graph-based node embeddings, \ie, $\mathbf{k},\mathbf{v}=\mathcal{H}(\mathbf{H})$, where $\mathbf{k},\mathbf{v} \in \mathbb{R}^{(N+1) \times D}$. Since each node is visited only once, RF applies mask $\mathbf{M}_{j}$ to the visited nodes for computing the attention weights, denoted as $\mathbf{A}_{j}=\texttt{Softmax}(\frac{\mathbf{q}_{j}\mathbf{k}^{\top}}{\sqrt{D}} \odot \mathbf{M}_{j})$, where $\mathbf{A}_{j}, \mathbf{M}_{j} \in \mathbb{R}^{N+1}$. The function $\texttt{Softmax}(\cdot)$ denotes the Softmax operation, and $\odot$ ensures that multiplication values for visited nodes are set to $-\infty$. The context query is computed as $\tilde{\mathbf{q}}_j=\mathcal{H}(\mathbf{A}_{j} \mathbf{v})$, and the candidate node representations are obtained as $\tilde{\mathbf{k}}=\mathcal{H}(\mathbf{H})$. The action probability distribution is derived as follows:
\begin{equation}
    \mathbf{D}_{j}= \tilde{\mathbf{q}}_j \tilde{\mathbf{k}}^{\top} /{\sqrt{D}},
    \label{eq: rl-dis}
\end{equation}
where $\tilde{\mathbf{q}}_j \in \mathbb{R}^{D}$, $\tilde{\mathbf{k}} \in \mathbb{R}^{(N+1) \times D}$, and $\mathbf{D}_{j} \in \mathbb{R}^{N+1}$. To generate solutions, the unnormalized log-probability (logit) is calculated as $\mathbf{u}_{j}= \xi \cdot \texttt{Tanh}(\mathbf{D}_{j}) \odot \mathbf{M}_{j}$, where $\texttt{Tanh}(\cdot)$ is the Hyperbolic Tangent operation, and $\xi=10$ is a predefined clipping hyperparameter. The final probability for selecting the $k$-th node is calculated by applying the Softmax function, denoted as $\mathbf{P}_{j}=\texttt{Softmax}(\mathbf{u}_{j})$.
If all constraints are satisfied, the node with the highest probability is selected as the next node to visit. Otherwise, the depot is selected.
\begin{figure*}[t]%
\centering

    \includegraphics[width=1.0 \textwidth]{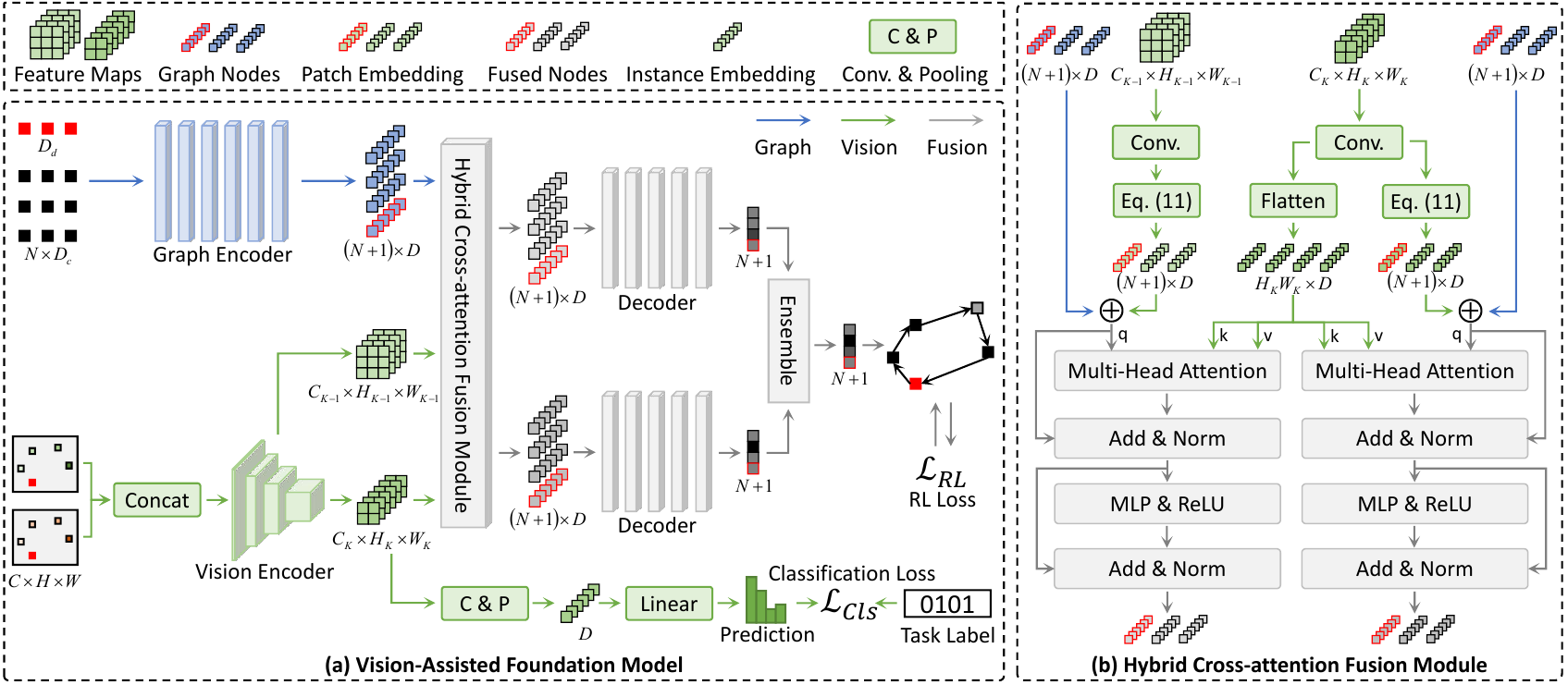}%
    \color{black}
    \caption{(a) In the graph modality, depot and customer attributes are processed by the graph encoder to generate node embeddings. In the vision modality, two images of each VRP instance are input into the vision encoder. The obtained feature maps are fused with the graph nodes to generate the final solutions, while a constraint classifier performs the constraint-aware auxiliary task. 
    \color{black}
    All modules are jointly trained with the weighted sum of $\mathcal{L}_{RL}$ and $\mathcal{L}_{Cls}$ (Eq.~\ref{eq: total-loss}).
    \color{black}
    (b) The hybrid fusion module uses the patch embeddings from the last two layers of the vision encoder, with a cross-attention mechanism applied to align the node-patch semantics.
    }
    \label{fig:framework}%
\end{figure*}
\subsubsection{Optimizing Objective}
\label{sec: RF-loss}
{To ensure minimal routing cost, RF employs the reinforcement learning (RL) algorithm to guide the training process.} This approach takes into account both the reward and the probability of selecting the current node. For the $i$-th trajectories, once the solutions $\{\pi_{i,1}, \pi_{i,1}, \dots, \pi_{i,N^{'}}\}$ are obtained, the reward is calculated by summing the node-to-node distances $r_i=\sum_{j=1}^{N^{'}-1} \| \mathbf{c}_{\pi_{i,j}}-\mathbf{c}_{\pi_{i,j+1}} \|_2$, where $\mathbf{c}=\{c^x, c^y\}$ represents the Euclidean coordinates, and $N^{'}$ denotes the total number of steps. The corresponding probabilities are represented by the likelihood $h_i=\sum_{j=1}^{N^{'}} \log(\mathbf{P}_{i,j,\pi_{i,j}})$, where $i$, $j$, and $\pi_{i,j}$ are the indexes for trajectory, step, and selected node, respectively. The training objective is
\begin{equation}
    \mathcal{L}_{RL}= \frac{1}{N} \sum_{i=1}^{N}(r_i-b)h_i,
\label{eq: rl}
\end{equation}
where $b$ is a shared baseline to reduce the variance of sampled gradients. It is calculated by averaging the reward from all trajectories, denoted as $b=\frac{1}{N} \sum_{i=1}^{N}r_i$.

\section{Methodology}
\subsection{Overall Workflow}
Fig.~\ref{fig:framework} (a) presents the overall workflow of our proposed VaFM, which integrates vision information into the graph-based neural VRP solver through a multi-modality framework. Both modalities use the coordinates, open route flag, duration limit, demand, and time window values as inputs. In the graph modality, these attributes are represented along the node axis and passed through a graph encoder. In the vision modality, two images are designed to represent all attributes of each individual VRP instance, which are then concatenated and fed into a vision encoder. The obtained vision patches are incorporated into the graph nodes using the proposed hybrid cross-attention fusion module. These fused embeddings are then passed to their respective decoders, which are ensembled to generate the final solutions. Additionally, VaFM incorporates a constraint classifier after the vision encoder to perform the auxiliary task. The overall training objective is defined as follows:
\begin{equation}
    \mathcal{L} =  \mathcal{L}_{RL} + \alpha \mathcal{L}_{Cls},
\label{eq: total-loss}
\end{equation}
where $\mathcal{L}_{RL}$ represents the RL loss for minimizing routing cost, and $\mathcal{L}_{Cls}$ denotes the classification loss for identifying constraint types. The coefficient $\alpha$ balances the loss terms and is empirically set to 0.1.

\subsection{Vision Input}
\color{black}
In the vision modality, we map customer coordinates as nodes and plot them into images. In multi-task scenarios, designing images for multiple attributes should follow:
\noindent\emph{Principle 1:} Represent attributes according to their type, using categorical cues for discrete variables and continuous cues for continuous variables.
\noindent\emph{Principle 2:} Use distinct visual channels for different attributes to prevent overlap, enabling a scalable combination of multiple attributes.

Based on these principles, each VRP instance is represented by demand and TW images, with depots shown as red nodes, L and O indicated by node shape and background color, and demand and time window conveyed through node color intensities. Below, we first introduce the vision input for CVRP and then extend it to cases with additional attributes.

\subsubsection{Vision Input for CVRP}
\label{sec: visInS}
\color{black}
The demand and TW images are denoted as $\mathbf{f}_1, \mathbf{f}_2 \in \mathbb{R}^{C \times H \times W}$, where $C$=3 denotes the number of channels, $H$ and $W$ represent the height and width of the images, respectively. Since the coordinates are within (0, 1), the corresponding image pixel index is calculated as follows:
\begin{equation}
    \begin{aligned}
        e_i^x=c_i^x \times (W-2M)+M, \\
        e_i^y=c_i^y \times (H-2M)+M.
    \end{aligned}
\label{eq: img_c}
\end{equation}
\color{black}
As shown in Fig.~\ref{fig: image}~(a), $M$ is a margin hyperparameter that prevents nodes from going beyond the image boundary. $(c_i^x, c_i^y)$ and $(e_i^x, e_i^y)$ represent the original Euclidean coordinates and the corresponding node pixel indexes, respectively. Each node is centered at $(e_i^x, e_i^y)$ and occupies an $S \times S$ pixel area, where larger-scale instances correspond to smaller $S$. 
\color{black}
The demand and TW images represent customers using green and white nodes, respectively, with the demand image indicating linehaul demand values through varying color intensities. 
\color{black}
Specifically, different instances may have widely varying demand levels, \eg, some instances having generally high demands and others low. To ensure sufficient color contrast within each image, the color intensity $s_i^l$ is normalized as follows:
\color{black}
\begin{equation}
    \begin{aligned}
        s_i^l=(\delta_i^l-\delta_{min})/(\delta_{max}-\delta_{min}),
    \end{aligned}
\end{equation}
where $\delta_{min}$, and $\delta_{max}$ denote the minimum and maximum demands among all customers within an instance.

\subsubsection{Extend CVRP to Tasks with a Single Attribute} 
\label{sec: visInP}
\begin{figure}[t]%
\color{black}
\centering
    \includegraphics[width=1.0 \columnwidth]{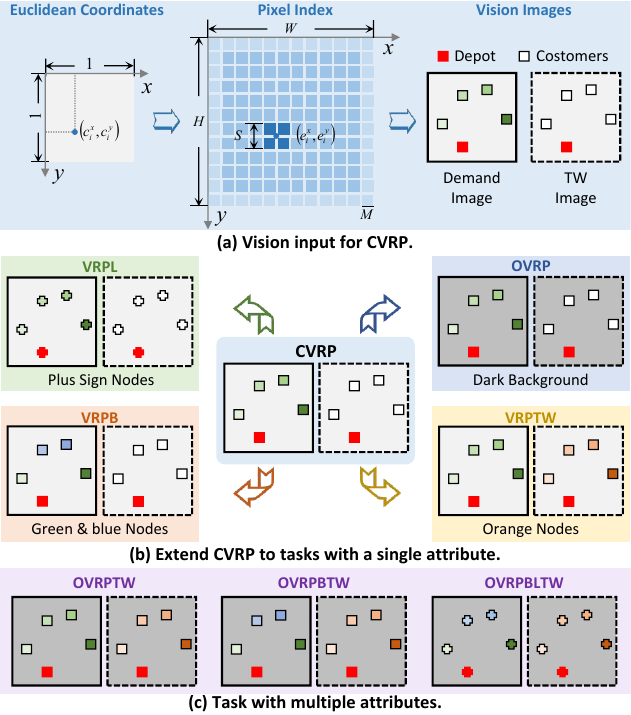}%
    \caption{(a) For CVRP, the Euclidean coordinates are converted into image pixel indices, which are then used to generate the demand and TW images. (b) CVRP is extended by adding one attribute. OVRP: background changes from light to dark grey; VRPL: nodes change from rectangles to plus signs. VRPB: demand image uses green/blue for linehaul and backhaul; VRPTW: orange nodes indicate time windows in the TW image. (c) We incorporate O-TW, O-TW-B, and O-TW-B-L into CVRP.
    }
    \label{fig: image}%
\end{figure}

To incorporate O into CVRP (\eg, OVRP), as shown in Fig.~\ref{fig: image}~(b), we invert the background pixel values from $P$ to $1-P$ (with $P$=0.25), resulting in a darker background. To incorporate L (\eg, VRPL), the depot and customer nodes are changed from rectangles to plus signs. These changes are applied to both images. In the VRPB, the demand image $\mathbf{f}_1$ uses green and blue nodes to represent linehaul and backhaul customers, respectively. The color intensity of each node is determined by its normalized demand:
\begin{equation}
    \begin{aligned}
        s_i^l=(\delta_i^l-\delta_{min})/(\delta_{max}-\delta_{min}), \\
        s_i^b=(\delta_i^b-\delta_{min})/(\delta_{max}-\delta_{min}),
    \end{aligned}
\end{equation}
where $s_i^l$ and $s_i^b$ represent the color intensities for linehaul and backhaul, respectively. For backhaul customers, $\delta_i^l = 0$, and for linehaul customers, $\delta_i^b = 0$. $\delta_{min}$, and $\delta_{max}$ denote the minimum and maximum values among all customers, \ie, $\{\delta_1^l+\delta_1^b, \delta_2^l+\delta_2^b, \dots, \delta_N^l+\delta_N^b\}$.
Similarly, in VRPTW, the TW image $\mathbf{f}_2$ uses orange nodes to represent time window information, with the color intensity defined as
\begin{equation}
    \begin{aligned}
        s_i^t=(t_i-t_{min})/(t_{max}-t_{min}),
    \end{aligned}
\end{equation}
where $t_i=t_i^e+t_i^s$ represents the sum of the early time and service time. The terms $t_{\text{min}}$ and $t_{\text{max}}$ denote the minimum and maximum values within $\{t_1^e+t_1^s, t_2^e+t_2^s, \dots, t_N^e+t_N^s\}$. 
\color{black}
We combine these constraint designs to formulate vision images for 16 VRP variants, with several examples shown in Fig.~\ref{fig: image}~(c).
\color{black}

\color{black}
\subsection{Vision Encoder}
To extract visual representations, we utilize a CNN-based vision encoder to process two images. Specifically, $\mathbf{f}_1$ and $\mathbf{f}_2$ are concatenated along the channel axis to form the encoder input $\mathbf{f} \in \mathbb{R}^{2C \times H \times W}$. It is subsequently passed through ResNet-18, which comprises an initial convolutional layer followed by various residual layers. We modify the input channel {number} of the initial layer from $C$ to $2C$ and maintain the subsequent layers' settings. The encoding process can be represented as
\begin{equation}
    \begin{aligned}
        \mathbf{F}^1, \mathbf{F}^2, \dots, \mathbf{F}^{K}=\mathcal{E}_v(\mathbf{f}),
    \end{aligned}
    \label{eq: VaFM-enc}
\end{equation}
where $\mathcal{E}_v(\cdot)$ is the CNN-based vision encoder that consists of $K$ residual layers, and $\mathbf{F}^k \in \mathbb{R}^{C_k \times H_k \times W_k}$ represents the encoded feature maps from the $k$-th layer. As the layer number increases, the number of channels doubles, while the height and width are halved, \ie, $C_{k+1}=2C_k$, $H_{k+1}=H_k/2$, and $W_{k+1}=W_k/2$. These feature maps are fused with the graph modality to generate solutions, followed by a constraint classifier after the encoder for the auxiliary task.

\subsection{Solution Generation}
The final solutions are derived from the multi-scale feature maps and graph-based node embeddings. Specifically, feature maps from the last two layers are integrated into node embeddings using the proposed hybrid cross-attention fusion module. This produces two fused embeddings, which are independently decoded to generate the final solutions. 
\color{black}
As the fusion process is similar for both layers, we use the $k$-th layer as a representative example in the following description to avoid redundancy.
\color{black}
\subsubsection{Hybrid Cross-Attention Fusion Module}
As described in Section~\ref{sec: RF-enc} and Eq.~\ref{eq: VaFM-enc}, the node embeddings and feature maps from the $k$-th layer are denoted as $\mathbf{H} \in \mathbb{R}^{(N+1) \times D}$ and $\mathbf{F}^k \in \mathbb{R}^{C_k \times H_k \times W_k}$, respectively. 
\color{black}
This module is used to fuse two embeddings into $\Tilde{\mathbf{E}}^k \in \mathbb{R}^{(N+1) \times D}$.

\noindent\textbf{Dimension Alignment for Patch Embeddings.} As illustrated in Fig.~\ref{fig:framework} (b), a $1 \times 1$ convolutional layer $\mathcal{C}(\cdot)$ performs a weighted linear transformation on the spatial feature map of size $H_k \times W_k$, thereby projecting the feature dimension from $C_k$ to $D$. Specifically, for each spatial location $(w, h)$, the output feature at the $d$-th channel is computed as
\begin{equation}
\Tilde{\mathbf{F}}^k_{d,w,h} = \sum_{c=1}^{C_k} W_{d,c}, \mathbf{F}^k_{c,w,h} + b_d,
\end{equation}
where $W \in \mathbb{R}^{D \times C_k}$ and $b \in \mathbb{R}^{D}$ are learnable weights and biases, respectively, resulting in $\Tilde{\mathbf{F}}^k \in \mathbb{R}^{D \times H_k \times W_k}$. 
\color{black}
Subsequently, the patch embedding for node $i$ is derived using the corresponding pixel indexes:
\begin{equation}
    \begin{aligned}
        \mathbf{P}_i^k = \Tilde{\mathbf{F}}_{w,h}^k, \quad w = \left\lfloor \frac{W_k}{W} e_i^x \right\rfloor, \quad h = \left\lfloor \frac{H_k}{H} e_i^y \right\rfloor,
    \end{aligned}
    \label{eq: patch_emb}
\end{equation}
where $(e_i^x, e_i^y)$ is calculated in Eq.~\ref{eq: img_c}. $\Tilde{\mathbf{F}}_{h,w}^k \in \mathbb{R}^{D}$ represents the features at the $h$-th row and $w$-th column of the corresponding feature map, where the floor function $\left\lfloor \cdot \right\rfloor$ ensures that $w$ and $h$ are integers.

\color{black}
\noindent\textbf{Graph-vision Integration.} Instead of directly adding graph and vision embeddings, we adopt a hybrid cross-attention mechanism to enable adaptive receptive fields. This mechanism combines the node embeddings with patch embeddings from different layers (corresponding to varying receptive fields), and further enhances representation ability by fusing the deepest-layer patch features. 
\color{black}
Specifically, queries are constructed by combining the node and patch embeddings:
\begin{equation}
    \begin{aligned}
        \mathbf{q}^k=\mathbf{H} + \beta \mathbf{P}^k, \ \
    \end{aligned}
    \label{eq: gra-vis}
\end{equation}
where $\mathbf{H}$ is the graph node embedding calculated in Section~\ref{sec: RF-enc}, the patch embedding $\mathbf{P}^k=\{\mathbf{P}_0^k,\mathbf{P}_1^k, \dots, \mathbf{P}_N^k\}$, both having the feature dimension $\mathbf{H}, \mathbf{P}^k \in \mathbb{R}^{(N+1) \times D}$. $\beta$ is the coefficient balancing graph and vision information, resulting in $\mathbf{q}^k \in \mathbb{R}^{(N+1) \times D}$.
The feature map from the final layer is flattened to serve as the key and value embeddings:
\begin{equation}
    \begin{aligned}
        \mathbf{k},\mathbf{v}=\mathcal{H}(\texttt{Flatten}(\Tilde{\mathbf{F}}^{K})), \ \
    \end{aligned}
\end{equation}
where $\mathbf{k}, \mathbf{v} \in \mathbb{R}^{H_{K} W_{K} \times D}$, $\texttt{Flatten}(\cdot)$ represents the flatten operation, and $\mathcal{H}(\cdot)$ is the linear layer to project key and value embeddings. 
The key and value embeddings are subsequently fused by the query embedding from the $k$-th layer of the vision encoder. It is implemented by using the transformer layer, with the attention matrices expressed as:
\begin{equation}
    \begin{aligned}
        \mathbf{A}^k=\mathbf{q}^k\mathbf{k}^{\top}/\sqrt{D},
    \end{aligned}
\end{equation}
{where $\mathbf{A}^k \in \mathbb{R}^{(N+1) \times H_k W_k}$. The fused node embedding is}
\begin{equation}
    \begin{aligned}
        {\mathbf{E}}^k=\texttt{IN}(\mathbf{q}^k+\mathcal{H}(\mathbf{A}^k \mathbf{v})), \ \
        \Tilde{\mathbf{E}}^k=\texttt{IN}({\mathbf{E}}^k+\texttt{MLP}({\mathbf{E}}^k)),
    \end{aligned}
\end{equation}
where $\mathbf{E}^k, \Tilde{\mathbf{E}}^k \in \mathbb{R}^{(N+1) \times D}$, $\texttt{IN}(\cdot)$ represents the instance normalization layer, and $\texttt{MLP}(\cdot)$ refers to the multi-layer perceptron (MLP) with a hidden dimension of $4D$. It consists of two linear layers, separated by a ReLU activation.

\subsubsection{Ensembled Decoder}
The fused embeddings from the last two layers are denoted as $\Tilde{\mathbf{E}}^{K-1}$ and $\Tilde{\mathbf{E}}^{K}$, respectively. 
\color{black}
At each decoding step $j$, the corresponding action probability distributions are represented by $\mathbf{D}_{j}^{K-1}$ and $\mathbf{D}_{j}^{K}$, as calculated in Eq.~\ref{eq: rl-dis}. The final probability distribution is ensembled by averaging decoder outputs, expressed as
\begin{equation}
    \begin{aligned}
        \mathbf{D}_j = \frac{1}{2}(\mathbf{D}_j^{K-1} + \mathbf{D}_j^{K}).
    \end{aligned}
\end{equation}
We follow the pipeline outlined in Section~\ref{sec: RF-dec} to compute the final probabilities for selecting the next node. Subsequently, we employ the RL loss $\mathcal{L}_{RL}$ from Section~\ref{sec: RF-loss} to minimize the routing cost.

\subsection{Constraint-aware Auxiliary Task}
As shown in Fig.~\ref{fig: image}, the CVRP and OVRP images exhibit significant background discrepancies. This may lead the model to overemphasize the features of O while neglecting the characteristics of L and B, which occupy fewer pixels. \textcolor{black}{To mitigate this issue, we introduce a constraint-aware auxiliary task to encourage more balanced learning across different constraints.
This task requires the model to identify the constraint type from the image, which can promote attention to constraints with lower pixel presence.}
As illustrated in Fig.~\ref{fig:framework} (a), the feature map from the final layer of the vision encoder is utilized to construct the instance-level embedding, denoted as $\mathbf{F}^{'}=\texttt{Pool}(\mathcal{C}(\mathbf{F}^{K}))$, where $\mathbf{F}^{'} \in \mathbb{R}^{D}$, $\texttt{Pool}(\cdot)$ and $\mathcal{C}(\cdot)$ represent the average pooling layer and a $1 \times 1$ convolutional layer, respectively. $\mathbf{F}^{'}$ is then fed into a constraint classifier to generate the prediction $\hat{\mathbf{T}}=\mathcal{H}(\mathbf{F}^{'})$, where $\hat{\mathbf{T}} \in \mathbb{R}^{4}$ represents the logits corresponding to the constraints O, L, B, and TW, and $\mathcal{H}(\cdot)$ denotes the classifier which consists of a linear layer. We employ the binary cross-entropy (BCE) loss~\cite{goodfellow2016deep} as the training objective:
\begin{equation}
    \mathcal{L}_{Cls} = - \frac{1}{\overline{C}} \sum_{i=1}^{\overline{C}} \left[ \mathbf{T}_i \log(\sigma(\hat{\mathbf{T}}_i)) + (1 - \mathbf{T}_i) \log(1 - \sigma(\hat{\mathbf{T}}_i)) \right],
\end{equation}
where $\overline{C}=4$ is the number of constraint classes, ${\mathbf{T}}_i$ is the corresponding one-hot label for the $i$-th constraint, and $\sigma(\cdot)$ is the sigmoid function. 
\color{black}
This part is jointly trained within VaFM, with the overall training objective defined in Eq.~\ref{eq: total-loss}.
\color{black}

%% file: 4-exp.tex
\section{Experiments}
\input{tables/1-SOTA-MTL}

\subsection{Experimental Setup}
\subsubsection{Dataset}
We evaluate the performance of our method using 16 VRP variants. Following~\cite{berto2024routefinder}, node locations are represented as two-dimensional Euclidean coordinates sampled from a uniform distribution within the range $[0,1)$. Each vehicle is initialized with a capacity of $Q=$1 at the depot.
The linehaul and backhaul nodes share the same demand configuration, where demands are uniformly distributed integer values in the range $[1,10)$, which is subsequently scaled by a factor of $30+N/5$, with $N$ representing the number of customers. 
% 
% The depot demand is set to 0. 
% 
For instances involving backhaul, 20\% of the customers are randomly selected and reassigned from linehaul to backhaul.
In the time window formulation, the vehicle speed is set to 1, and the {early} times are randomly selected from the range $[0.0126, 4.25]$. The service times and window lengths are drawn from the ranges $[0, 0.15)$ and $[1.8, 2.0)$, respectively. The {late} times are computed by adding the {early} times and window lengths. {For instances without TW, all time values are set to zero.}
The distance limit is set to $l$=3 for instances with L, and instances without L are assigned a value of zero. A total of 100,000 instances are uniformly generated across 16 variants for training. During evaluation, each task employs 128 instances for validation and 1,000 for testing. {In this paper, we compare the performance of VaFM with the state-of-the-art (SOTA) methods by training models on both $N$=50 and $N$=100, while mainly using $N$=50 for ablation, sensitivity, and other analysis studies.}
\input{tables/1-sota-gain}

\subsubsection{Implementation Details}
Our method is developed using PyTorch~\cite{NEURIPS2019_9015}. The training and evaluation experiments are conducted on an AMD EPYC 7702P 24-core CPU with a single RTX 3090 GPU.
% 

% 
% Graph-based and vision-based inputs share location information and four constraints (\ie, O, L, B, and TW). All infinite values are replaced with zero.
% Demands are divided into linehaul and backhaul, while time windows are defined by the early time, last time, and service time. 
% 
In the graph encoder, we adopt the one from RF-MoE-L~\cite{berto2024routefinder}, which consists of four experts and hierarchical gating. The embedding dimension $D$ is set to 128.
We employ the  ResNet-18  pre-trained on ImageNet ~\cite{he2016deep} as the vision encoder, consisting of $K$=4 residual layers. The input image size is $H$=$W$=224, and the initial convolutional layer's channel number is modified from 3 to 6. Feature maps are extracted from the last two layers, yielding dimensions $C_3$=256, $C_4$=512, and map sizes $H_3$=$W_3$=14, $H_4$=$W_4$=7. In the constraint classifier, we use a $1 \times 1$ convolutional kernel with a stride of 1.
In the decoding process, we continue to use the RF-MoE-L decoder with the same feature dimension of 128. All modules are jointly trained by using the Adam optimizer. 
\color{black}
It is initialized with a learning rate of $3 \times 10^{-4}$ and a weight decay of $10^{-6}$. 
\color{black}
During the 300 training epochs, the learning rate is reduced using the MultiStepLR strategy by a factor of 0.1 at the 270th and 295th epochs. The checkpoint with the best validation performance is used for testing. 
\color{black}
\textbf{Code is available at: \url{https://github.com/gshuangchun/VaFM}}.
\color{black}

\subsubsection{Evaluation Metrics}
During the inference stage, we generate greedy solutions using multi-starts and eight symmetric dihedral augmentations from~\cite{kwon2020pomo}, resulting in $8 \times N$ solutions per instance. The model produces these solutions independently of the constraint classifier. To assess the model's performance, we use two widely adopted metrics in VRPs: the total routing length ("Obj.") and the performance gap ("Gap") to the strong baseline, HGS-PyVRP~\cite{wouda2024pyvrp}. Additionally, we report the inference time (latency). All metrics are computed over 1,000 test instances, where "Obj." and "Gap" represent the average values per instance, and the inference time reflects the total computation cost for {all 1,000 test instances}. 
\subsection{Comparison with the State-of-the-Arts}
% \subsubsection{Multi-task VRPs}
\subsubsection{Comparison Across Different Tasks}

We compare VaFM with the traditional and SOTA neural solvers. In traditional solvers, HGS-PyVRP extends the SOTA heuristic algorithm HGS-CVRP on 16 VRPs, while OR-Tools is a commonly used method in solving various VRP variants through constraint programming. {The implementation follows the setup as illustrated in \cite{berto2024routefinder}. Both methods adopt time limits of 10 and 20 seconds for instances with 50 and 100 nodes, respectively.}
Neural solvers include MTPOMO~\cite{liu2024multi}, the first multi-task VRP baseline, and MVMoE~\cite{zhou2024mvmoe}, which utilizes mixture-of-experts~\cite{fedus2022review} to enhance model performance. Additionally, we incorporate two RF variants~\cite{berto2024routefinder}: RF-POMO, derived from MTPOMO, and RF-MoE-L based on MVMoE-L. Compared to MVMoE, MVMoE-L incorporates a hierarchical gating mechanism. Accordingly, we develop two VaFM variants: VaFM-POMO and VaFM-MoE-L.
Table~\ref{tab: sota} presents the total routing length, {gap to strong baseline}, and inference time on test instances with 50 and 100 customer nodes.
The results demonstrate that VaFM-MoE-L achieves superior performance across most VRP variants, with the best results in 15 out of 16 cases for $N$=50 and 12 out of 16 cases for $N$=100. Compared to RF-MoE-L, VaFM-MoE-L demonstrates gap improvements of 0.542\% and 0.427\% in OVRPB with $N$=50 and OVRPBL with $N$=100, respectively. 
\color{black}
Moreover, we observe that VaFM suggests modest benefits in CVRP, with VaFM-MoE-L narrowing a 0.028\% gap from RF-MoE-L in $N$=50. This may be attributed to redundancy in visual inputs. For example, the visual input includes both demand and TW images, but the TW image provides little useful information for CVRP and may even introduce noise.
\color{black}
Regarding the inference time, all neural solvers demonstrate superiority over conventional HGS-PyVRP and OR-Tools. {However, VaFM-MoE-L takes slightly longer than RF-MoE-L when solving each VRP variant. For variants with $N$=100,} RF-MoE-L and VaFM-MoE-L require 12–14s and 18-22s for inference, respectively.
\subsubsection{Comparison Across Different Constraint Numbers}
We investigate the improvements of VaFM over the SOTA method RF across tasks with varying numbers of constraints. Based on CVRP, we present the performance across various combinations of constraints. {Table~\ref{tab: sota-gain} presents the values of "Obj." and "Gap", as well as their improvements from RF (RF-MoE-L in Table~\ref{tab: sota}) to VaFM (VaFM-MoE-L in Table~\ref{tab: sota}). We observe that variants with different constraint numbers exhibit similar averaged values of "Obj." and "Gap". For RF with $N$=50, the averaged "Gap" values for tasks with 0, 1, 2, 3, and 4 constraints are distributed as \{1.332\%, 2.343\%, 2.496\%, 1.766\%, 1.259\%\}. This ensures that any observed performance improvements are not attributable to inherently larger "Gap" values in multi-constraint tasks. In contrast, the "($\Delta$)" results indicate that VaFM achieves greater improvements in tasks with multiple constraints. Specifically, VaFM outperforms RF by a 0.028\% {"Gap"} in CVRP with $N$=50. As the constraint number increases, VaFM achieves {average "Gap" improvements} of 0.256\%, 0.362\%, and 0.354\% in tasks with 1, 2, and 3 constraints, respectively. These results highlight VaFM’s superiority in handling tasks with complex constraints, where the improvements are attributed to the model’s effectiveness.}
\begin{figure*}[t]%
\color{black}
\centering
    \includegraphics[width=1 \textwidth]{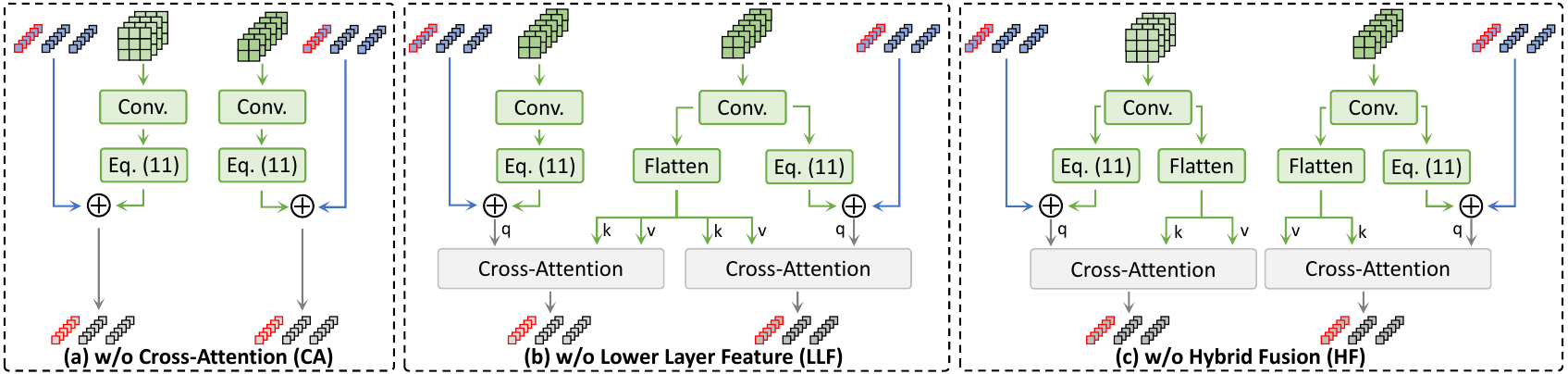}%
    \caption{{Ablation study on the hybrid cross-attention fusion module:
(a) directly adds the graph and patch embeddings;
(b) uses the same patch embedding from the last encoder layer;
(c) employs the $\mathbf{q}$, $\mathbf{k}$, and $\mathbf{v}$ from the patch embedding within the same encoder layer.}
    }
    \label{fig: exp-fuse}%
    \vspace{-8pt}
\end{figure*}

\subsection{Ablation Studies}
\subsubsection{Ablation on Multi-task VRP Image}
To evaluate the effectiveness of the proposed vision input, we conduct ablation studies by systematically removing each design for the constraint. 
O \ding{55} means the background design is removed, leaving a uniform color. B \ding{55} indicates that backhaul shares the same node color as linehaul. L \ding{55} denotes that tasks with or without L have the same node shape. TW \ding{55} refers to using an image with only the first demand information, excluding the second TW image. Additionally, we design an image that simultaneously removes all constraints, denoted as O \ding{55}, B \ding{55}, L \ding{55}, and TW \ding{55} (last row).
The results in Table~\ref{tab: abl-att} show that incorporating any constraint design in the image improves performance. Specifically, removing B and TW leads to a more {severe performance drop} than removing O and L. For example, eliminating B causes a 0.175\% {performance drop}, while removing L results in a smaller {magnitude} of 0.008\%. These findings suggest that node-based constraints contribute more to {performance improvements} than instance-based ones.

\input{tables/3-Abl-img}

\input{tables/3-Abl-fusion}
\begin{figure}[t]%
\color{black}
\centering
    \includegraphics[width=1 \columnwidth]{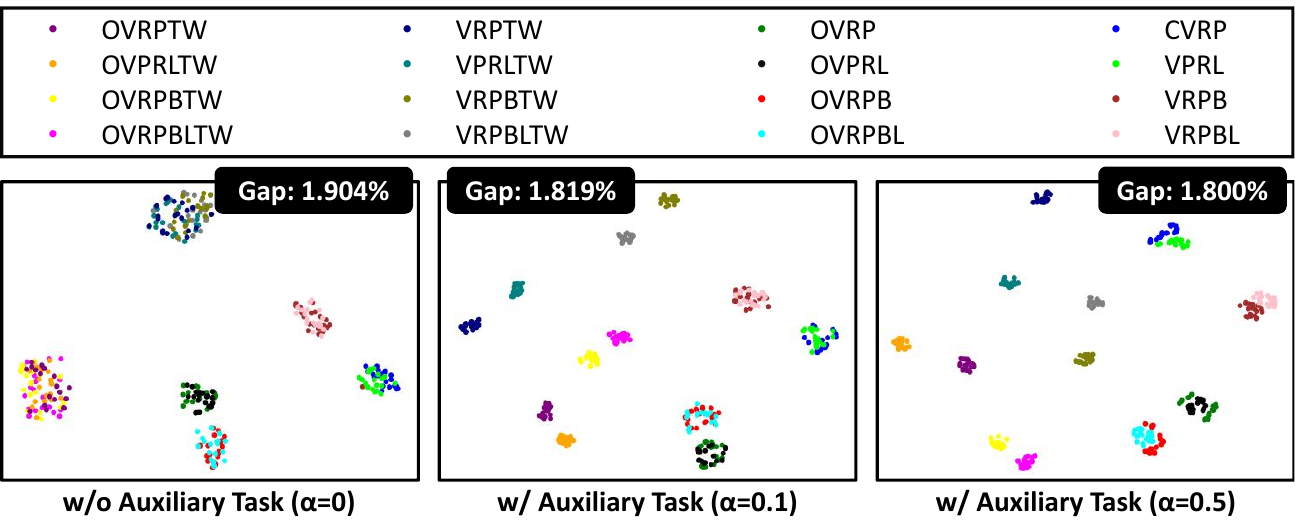}%
    \caption{{Variant distribution of VaFM under different loss weights.}
    }
    \label{fig: dis-task}%
    \vspace{-8pt}
\end{figure}
\input{tables/4-Abl-scale}

\color{black}
\subsubsection{Ablation on Hybrid Cross-attention Fusion Module}

To evaluate the effectiveness of each component within the hybrid cross-attention fusion module, we assess the model performance by individually removing the cross-attention mechanism (CA), the lower-layer feature (LLF), and the hybrid fusion (HF). As illustrated in Fig.~\ref{fig: exp-fuse}, removing CA directly uses the summed graph and patch features as the fused node embedding. Removing LLF retains only the patch embedding from the last encoder layer, while removing HF employs the $\mathbf{q}$, $\mathbf{k}$, and $\mathbf{v}$ from the same layer.
As shown in Table~\ref{tab: abl-key}, removing each module increases the Gap by 0.057\%, 0.056\%, and 0.068\% for CA, LLF, and HF, respectively.
\color{black}

\color{black}
\subsection{Impact of Auxiliary Task}
We further analyze the effectiveness of the constraint-aware auxiliary task by examining its impact on the performance gap and the resulting task distribution. As shown in Fig.~\ref{fig: dis-task}, “w/o Auxiliary Task ($\alpha$=0)” removes the constraint-classifier branch and excludes the classification loss from the training objective. The other two settings evaluate VaFM under different balance coefficients in Eq.~\ref{eq: total-loss}. The results indicate that our VaFM incorporating auxiliary task with $\alpha$=0.1 improves performance by 0.085\% in "Gap".
Regarding the task (variant) distribution, we extract feature maps from the last layer of the vision encoder and apply an adaptive average pooling layer to compute instance-level high-dimensional features. Subsequently, 20 instances are randomly selected from each variant, and T-SNE~\cite{van2008visualizing} is applied to project the feature space into 2D. 
As shown in Fig.~\ref{fig: dis-task}, VaFM without the auxiliary task exhibits a large feature distance between \{OVRPBLTW, OVRPBTW, OVRPLTW, OVRPTW\} and \{VRPBLTW, VRPBTW, VRPLTW, VRPTW\}. This suggests that the model overemphasizes constraints represented by different backgrounds (the image design of O) while overlooking those with minimal visual presence, such as B and L. In contrast, VaFM correctly distinguishes all eight TW variants, indicating that the auxiliary task enables the model to capture the semantics of O, B, and L in this setting. For tasks without TW, the auxiliary task brings modest improvements in recognizing L. We attribute this to the small weight assigned to the classification loss, and a larger balance coefficient (\eg, $\alpha$=0.5) can alleviate this. \textcolor{black}{This indicates that the recognition of B and L could be influenced by the auxiliary loss weight, as the classification loss serves only as additional supervision rather than directly optimizing VRP. More robust recognition may require redesigned visual encoding schemes, which we plan to explore in the future (\eg, using a striped background for L and striped nodes for B).}
\color{black}

\subsection{Impact of Patch Embeddings from Different Layers}
We continue to evaluate the effectiveness of patch embeddings from different layers of the vision encoder. We begin by establishing baseline models with varying numbers of decoders, each sharing input node embeddings from the graph encoder. {"Baseline$\times$1" refers to the reproduced results of RF-MoE-L from Table~\ref{tab: sota}, using the code provided by~\cite{berto2024routefinder}.} Subsequently, we incorporate patch embeddings from different layer combinations and compare their performance with the corresponding baselines. 
\color{black}
As shown in Table~\ref{tab: abl-scale}, when using a single layer, only the embeddings from layers 3 and 4 reduce the optimality gap compared to “Baseline$\times$1”, with deeper layers yielding better performance. In multi-layer integration, all combinations achieve smaller gaps than their corresponding baselines. In this paper, we adopt the second-best setting (\ie, layers 3 and 4) to balance performance and model complexity.
\color{black}

\input{tables/5-Dis-Gen-200}
\begin{figure}[t]%
\centering
    \includegraphics[width=1 \columnwidth]{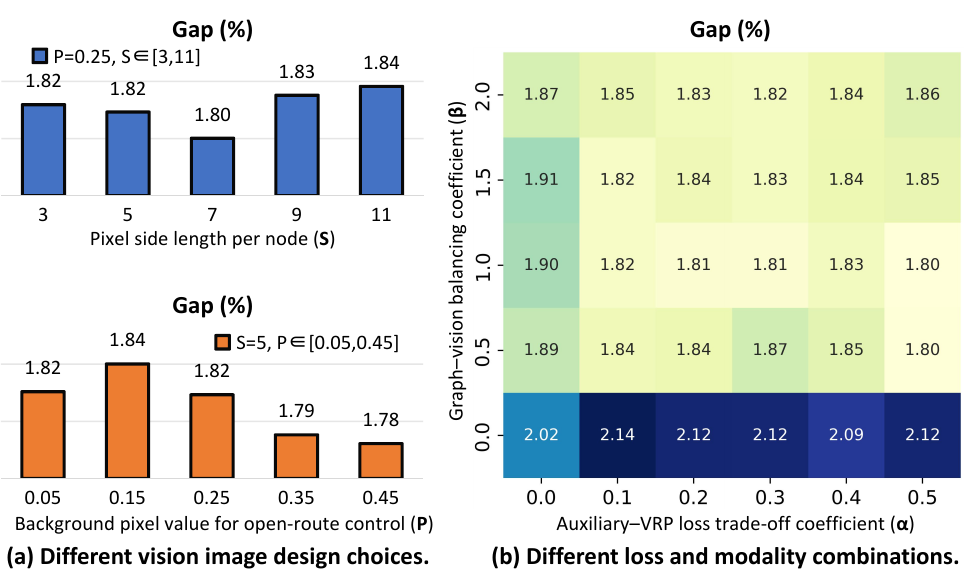}%
    \caption{Sensitivity study on image parameters and balancing coefficients.
    }
    \label{fig: alpha}%
    \vspace{-8pt}
\end{figure}

\subsection{Sensitivity Studies}
\color{black}
\subsubsection{Impact of Vision Image Design Parameters} In Section~\ref{sec: visInS}, $S$ denotes the pixel side length of each node, whereas in Section~\ref{sec: visInP}, $P$ represents the background pixel value of an image without open route (O). To investigate the impact of design choices, we evaluate VaFM with $P$=0.25 and $S \in \{3, 5, 7, 9, 11\}$, as well as with $S$=5 and $P \in \{0.05, 0.15, 0.25, 0.35, 0.45\}$. Fig.~\ref{fig: alpha}~(a) presents the averaged gap across 16 tasks under these settings. The results show that the gap increases when the pixel side length is too large or too small, and it reaches the minimum at $S$=7. For the background design, increasing $P$ leads to better performance. This suggests that a smaller color intensity difference between the backgrounds with and without $L$ is more effective. In this paper, we empirically set $S$=5 for $N$=50 and $S$=3 for $N$=100, while $P$=0.25 is used for both cases.

\subsubsection{Impact of Modality and Loss Balancing Coefficients}
\color{black}
In Eq.~\ref{eq: total-loss}, $\alpha$ adjusts the trade-off between the auxiliary task and the VRP objective, while in Eq.~\ref{eq: gra-vis}, $\beta$ serves as a coefficient that balances graph and vision modality information. To study the impact of different combinations on the optimization performance, we evaluate VaFM using different $\alpha$ and $\beta$ settings. All 30 combinations with $\alpha \in$ \{0.0, 0.1, 0.2, 0.3, 0.4, 0.5\} and $\beta \in$ \{0.0, 0.5, 1.0, 1.5, 2.0\} are examined across 16 task variants with 50 customer nodes. Fig.~\ref{fig: alpha} presents the {"Gap"} achieved under different combinations. The results show that VaFM consistently performs well with $\alpha \in [0.1, 0.5]$ and $\beta$=1.0. However, when $\alpha$ or $\beta$ is set to zero, {performance decreases significantly}, highlighting the effectiveness of the constraint-aware auxiliary task and the vision modality. In this paper, we {empirically} set $\alpha$=0.1 and $\beta$=1.0 for simplicity.

%% file: tables/1-SOTA-MTL.tex
\begin{table*}[htbp]
  \caption{Performance on multi-task VRPs. * denotes the strong baseline used to compute the gap. Best neural approach in \textbf{bold}; second \underline{underlined}.}
  \vspace{-4pt}
  \label{tab: sota}
  \begin{center}
  % % \begin{small}
  \renewcommand\arraystretch{1.05}  % 0.97
  \resizebox{0.98\textwidth}{!}{ 
  \begin{tabular}{ll|rrrrrr|ll|rrrrrr}
    \toprule
    % \midrule
    \multicolumn{2}{c|}{\multirow{2}{*}{Solver}} & \multicolumn{3}{c}{{$N=50$}} & \multicolumn{3}{c|}{$N=100$} & \multicolumn{2}{c|}{\multirow{2}{*}{Solver}} &
    \multicolumn{3}{c}{{$N=50$}} & \multicolumn{3}{c}{$N=100$} \\
    \cmidrule(lr){3-5} \cmidrule(lr){6-8} \cmidrule(lr){11-13} \cmidrule(lr){14-16}
     & & Obj. $\downarrow$ & Gap $\downarrow$ & Time $\downarrow$ & Obj. $\downarrow$ & Gap $\downarrow$ & Time $\downarrow$ & & & Obj. $\downarrow$ & Gap $\downarrow$ & Time & Obj. $\downarrow$ & Gap $\downarrow$ & Time $\downarrow$ \\
     
    \midrule
\multirow{8}*{\rotatebox{90}{CVRP}} & HGS-PyVRP & 10.287 & * & 4.6m & 15.543 & * & 9.2m & \multirow{8}*{\rotatebox{90}{OVRP}} & HGS-PyVRP & 6.494 & * & 4.6m & 9.730 & * & 9.2m \\
 & OR-Tools & 10.523 & 2.294\% & 4.6m & 16.361 & 5.263\% & 9.2m && OR-Tools & 6.555 & 0.939\% & 4.6m & 10.081 & 3.607\% & 9.2m \\
 & MTPOMO & 10.458 & 1.662\% & 2s & 15.796 & 1.628\% & 10s && MTPOMO & 6.818 & 4.989\% & 2s & 10.239 & 5.231\% & 10s \\
 & MVMoE & \textbf{10.414} & \textbf{1.235\%} & 3s & \textbf{15.759} & \textbf{1.390\%} & 13s && MVMoE & 6.760 & 4.096\% & 3s & 10.195 & 4.779\% & 13s \\
 & MVMoE-L & 10.448 & 1.565\% & 3s & \underline{15.777} & \underline{1.506\%} & 12s && MVMoE-L & 6.796 & 4.650\% & 2s & 10.224 & 5.077\% & 12s \\
 & RF-POMO & 10.438 & 1.468\% & 2s & 15.836 & 1.885\% & 10s && RF-POMO & 6.706 & 3.265\% & 2s & 10.204 & 4.872\% & 10s \\
 & RF-MoE-L & 10.424 & 1.332\% & 2s & 15.818 & 1.769\% & 12s && RF-MoE-L & 6.682 & 2.895\% & 2s & 10.165 & 4.471\% & 12s \\
 & VaFM-POMO & 10.432 & 1.412\% & 5s  & 15.812 & 1.730\% & 13s  && VaFM-POMO & \underline{6.677} & \underline{2.811\%} & 4s  & \underline{10.144} & \underline{4.258\%} & 13s \\
 & VaFM-MoE-L & \underline{10.421} & \underline{1.304\%} & 5s  & 15.802 & 1.665\% & 18s  && VaFM-MoE-L & \textbf{6.671} & \textbf{2.719\%} & 5s  & \textbf{10.141} & \textbf{4.228\%} & 19s \\

\midrule
\multirow{8}*{\rotatebox{90}{VRPB}} & HGS-PyVRP & 9.688 & * & 4.6m & 14.386 & * & 9.2m & \multirow{8}*{\rotatebox{90}{OVRPB}} & HGS-PyVRP & 6.897 & * & 4.6m & 10.304 & * & 9.2m \\
 & OR-Tools & 9.829 & 1.455\% & 4.6m & 15.010 & 4.338\% & 9.2m && OR-Tools & 6.940 & 0.623\% & 4.6m & 10.611 & 2.979\% & 9.2m \\
 & MTPOMO & 10.105 & 4.304\% & 2s & 15.012 & 4.351\% & 10s && MTPOMO & 7.269 & 5.394\% & 2s & 10.901 & 5.794\% & 10s \\
 & MVMoE & 10.009 & 3.313\% & 3s & \textbf{14.948} & \textbf{3.907\%} & 13s && MVMoE & 7.179 & 4.089\% & 3s & 10.846 & 5.260\% & 13s \\
 & MVMoE-L & 10.069 & 3.933\% & 2s & 14.983 & 4.150\% & 11s && MVMoE-L & 7.239 & 4.959\% & 2s & 10.874 & 5.532\% & 12s \\
 & RF-POMO & 10.012 & 3.344\% & 2s & 15.020 & 4.407\% & 10s && RF-POMO & 7.108 & 3.059\% & 2s & 10.816 & 4.969\% & 10s \\
 & RF-MoE-L & 9.977 & 2.983\% & 2s & 14.980 & 4.129\% & 11s && RF-MoE-L & 7.091 & 2.813\% & 2s & 10.773 & 4.552\% & 12s \\
 & VaFM-POMO & \underline{9.966} & \underline{2.874\%} & 3s  & 14.967 & 4.038\% & 13s  && VaFM-POMO & \underline{7.068} & \underline{2.472\%} & 4s  & \underline{10.745} & \underline{4.281\%} & 13s \\
 & VaFM-MoE-L & \textbf{9.948} & \textbf{2.689\%} & 5s  & \underline{14.956} & \underline{3.964\%} & 18s  && VaFM-MoE-L & \textbf{7.054} & \textbf{2.271\%} & 5s  & \textbf{10.737} & \textbf{4.197\%} & 19s \\

\midrule
\multirow{8}*{\rotatebox{90}{VRPBL}} & HGS-PyVRP & 9.688 & * & 4.6m & 14.373 & * & 9.2m & \multirow{8}*{\rotatebox{90}{OVRPBL}} & HGS-PyVRP & 6.904 & * & 4.6m & 10.310 & * & 9.2m \\
 & OR-Tools & 9.820 & 1.363\% & 4.6m & 15.084 & 4.947\% & 9.2m && OR-Tools & 6.949 & 0.652\% & 4.6m & 10.613 & 2.939\% & 9.2m \\
 & MTPOMO & 10.112 & 4.377\% & 2s & 15.023 & 4.522\% & 10s && MTPOMO & 7.274 & 5.359\% & 2s & 10.903 & 5.752\% & 10s \\
 & MVMoE & 10.018 & 3.406\% & 3s & \textbf{14.951} & \textbf{4.021\%} & 13s && MVMoE & 7.191 & 4.157\% & 3s & 10.858 & 5.315\% & 13s \\
 & MVMoE-L & 10.080 & 4.046\% & 2s & 14.993 & 4.314\% & 11s && MVMoE-L & 7.239 & 4.852\% & 2s & 10.886 & 5.587\% & 12s \\
 & RF-POMO & 10.026 & 3.489\% & 2s & 15.030 & 4.571\% & 10s && RF-POMO & 7.117 & 3.085\% & 2s & 10.825 & 4.995\% & 10s \\
 & RF-MoE-L & 9.992 & 3.138\% & 2s & 14.982 & 4.237\% & 12s && RF-MoE-L & 7.099 & 2.824\% & 2s & 10.781 & 4.568\% & 12s \\
 & VaFM-POMO & \underline{9.976} & \underline{2.969\%} & 3s  & 14.960 & 4.086\% & 13s  && VaFM-POMO & \underline{7.078} & \underline{2.519\%} & 4s  & \underline{10.757} & \underline{4.334\%} & 13s \\
 & VaFM-MoE-L & \textbf{9.959} & \textbf{2.798\%} & 5s  & \underline{14.958} & \underline{4.072\%} & 18s  && VaFM-MoE-L & \textbf{7.062} & \textbf{2.291\%} & 5s  & \textbf{10.737} & \textbf{4.141\%} & 19s \\

\midrule
\multirow{8}*{\rotatebox{90}{VRPBLTW}} & HGS-PyVRP & 18.361 & * & 4.6m & 29.026 & * & 9.2m & \multirow{8}*{\rotatebox{90}{OVRPBLTW}} & HGS-PyVRP & 11.597 & * & 4.6m & 19.005 & * & 9.2m \\
 & OR-Tools & 18.422 & 0.332\% & 4.6m & 29.830 & 2.770\% & 9.2m && OR-Tools & 11.612 & 0.129\% & 4.6m & 19.198 & 1.016\% & 9.2m \\
 & MTPOMO & 18.841 & 2.614\% & 2s & 30.232 & 4.155\% & 11s && MTPOMO & 11.963 & 3.156\% & 2s & 19.626 & 3.268\% & 11s \\
 & MVMoE & 18.715 & 1.928\% & 3s & 30.216 & 4.100\% & 15s && MVMoE & 11.847 & 2.156\% & 3s & 19.588 & 3.068\% & 15s \\
 & MVMoE-L & 18.773 & 2.244\% & 3s & 30.223 & 4.124\% & 13s && MVMoE-L & 11.883 & 2.466\% & 3s & 19.605 & 3.157\% & 14s \\
 & RF-POMO & 18.628 & 1.454\% & 2s & 30.094 & 3.679\% & 11s && RF-POMO & 11.735 & 1.190\% & 2s & 19.429 & 2.231\% & 11s \\
 & RF-MoE-L & 18.622 & 1.421\% & 3s & 30.060 & 3.562\% & 13s && RF-MoE-L & 11.743 & 1.259\% & 3s & 19.402 & 2.089\% & 14s \\
 & VaFM-POMO & \underline{18.610} & \underline{1.356\%} & 4s  & \underline{29.990} & \underline{3.322\%} & 14s  && VaFM-POMO & \underline{11.727} & \underline{1.122\%} & 4s  & \underline{19.365} & \underline{1.895\%} & 15s \\
 & VaFM-MoE-L & \textbf{18.572} & \textbf{1.150\%} & 5s  & \textbf{29.985} & \textbf{3.305\%} & 21s  && VaFM-MoE-L & \textbf{11.712} & \textbf{0.991\%} & 5s  & \textbf{19.359} & \textbf{1.861\%} & 22s \\

\midrule
\multirow{8}*{\rotatebox{90}{VRPBTW}} & HGS-PyVRP & 18.167 & * & 4.6m & 29.000 & * & 9.2m & \multirow{8}*{\rotatebox{90}{OVRPBTW}} & HGS-PyVRP & 11.590 & * & 4.6m & 19.167 & * & 9.2m \\
 & OR-Tools & 18.374 & 1.139\% & 4.6m & 29.964 & 3.324\% & 9.2m && OR-Tools & 11.610 & 0.173\% & 4.6m & 19.314 & 0.767\% & 9.2m \\
 & MTPOMO & 18.797 & 3.468\% & 2s & 30.325 & 4.569\% & 11s && MTPOMO & 11.957 & 3.167\% & 2s & 19.780 & 3.198\% & 11s \\
 & MVMoE & 18.684 & 2.846\% & 3s & 30.319 & 4.548\% & 15s && MVMoE & 11.849 & 2.235\% & 3s & 19.752 & 3.052\% & 15s \\
 & MVMoE-L & 18.760 & 3.264\% & 2s & 30.323 & 4.562\% & 14s && MVMoE-L & 11.880 & 2.502\% & 2s & 19.770 & 3.146\% & 14s \\
 & RF-POMO & 18.589 & 2.323\% & 2s & 30.181 & 4.072\% & 11s && RF-POMO & 11.733 & 1.234\% & 2s & 19.579 & 2.150\% & 11s \\
 & RF-MoE-L & 18.585 & 2.301\% & 3s & 30.157 & 3.990\% & 13s && RF-MoE-L & 11.737 & 1.268\% & 3s & 19.558 & 2.040\% & 14s \\
 & VaFM-POMO & \underline{18.578} & \underline{2.262\%} & 4s  & \underline{30.099} & \underline{3.790\%} & 14s  && VaFM-POMO & \underline{11.720} & \underline{1.123\%} & 4s  & \underline{19.519} & \underline{1.834\%} & 15s \\
 & VaFM-MoE-L & \textbf{18.536} & \textbf{2.031\%} & 5s  & \textbf{30.090} & \textbf{3.758\%} & 20s  && VaFM-MoE-L & \textbf{11.703} & \textbf{0.974\%} & 6s  & \textbf{19.511} & \textbf{1.795\%} & 22s \\

\midrule
\multirow{8}*{\rotatebox{90}{VRPL}} & HGS-PyVRP & 10.328 & * & 4.6m & 15.637 & * & 9.2m & \multirow{8}*{\rotatebox{90}{OVRPL}} & HGS-PyVRP & 6.510 & * & 4.6m & 9.709 & * & 9.2m \\
 & OR-Tools & 10.570 & 2.343\% & 4.6m & 16.466 & 5.302\% & 9.2m && OR-Tools & 6.571 & 0.937\% & 4.6m & 10.047 & 3.481\% & 9.2m \\
 & MTPOMO & 10.502 & 1.685\% & 2s & 15.905 & 1.714\% & 12s && MTPOMO & 6.839 & 5.054\% & 2s & 10.210 & 5.160\% & 10s \\
 & MVMoE & 10.457 & 1.249\% & 3s & 15.865 & 1.458\% & 13s && MVMoE & 6.777 & 4.101\% & 3s & 10.169 & 4.738\% & 13s \\
 & MVMoE-L & 10.488 & 1.549\% & 2s & 15.885 & 1.586\% & 10s && MVMoE-L & 6.803 & 4.501\% & 2s & 10.199 & 5.047\% & 12s \\
 & RF-POMO & 10.482 & 1.491\% & 2s & 15.955 & 2.034\% & 10s && RF-POMO & 6.727 & 3.333\% & 2s & 10.176 & 4.810\% & 10s \\
 & RF-MoE-L & 10.464 & 1.317\% & 2s & 15.924 & 1.835\% & 12s && RF-MoE-L & 6.700 & 2.919\% & 2s & \underline{10.135} & \underline{4.388\%} & 12s \\
 & VaFM-POMO & \underline{10.448} & \underline{1.165\%} & 3s  & \underline{15.926} & \underline{1.850\%} & 13s  && VaFM-POMO & \underline{6.686} & \underline{2.708\%} & 4s  & 10.144 & 4.485\% & 13s \\
 & VaFM-MoE-L & \textbf{10.436} & \textbf{1.050\%} & 5s  & \textbf{15.921} & \textbf{1.818\%} & 19s  && VaFM-MoE-L & \textbf{6.676} & \textbf{2.548\%} & 5s  & \textbf{10.134} & \textbf{4.376\%} & 19s \\

\midrule
\multirow{8}*{\rotatebox{90}{VRPLTW}} & HGS-PyVRP & 15.951 & * & 4.6m & 25.678 & * & 9.2m & \multirow{8}*{\rotatebox{90}{OVRPLTW}} & HGS-PyVRP & 10.455 & * & 4.6m & 16.962 & * & 9.2m \\
 & OR-Tools & 16.036 & 0.533\% & 4.6m & 26.156 & 1.862\% & 9.2m && OR-Tools & 10.465 & 0.096\% & 4.6m & 17.100 & 0.814\% & 9.2m \\
 & MTPOMO & 16.480 & 3.316\% & 2s & 26.684 & 3.918\% & 11s && MTPOMO & 10.803 & 3.329\% & 2s & 17.589 & 3.696\% & 11s \\
 & MVMoE & 16.368 & 2.614\% & 3s & 26.655 & 3.805\% & 14s && MVMoE & 10.718 & 2.516\% & 3s & 17.553 & 3.484\% & 15s \\
 & MVMoE-L & 16.437 & 3.047\% & 3s & 26.674 & 3.879\% & 13s && MVMoE-L & 10.750 & 2.822\% & 3s & 17.577 & 3.626\% & 14s \\
 & RF-POMO & 16.303 & 2.207\% & 2s & 26.572 & 3.482\% & 11s && RF-POMO & 10.611 & 1.492\% & 2s & 17.429 & 2.753\% & 11s \\
 & RF-MoE-L & 16.298 & 2.175\% & 3s & 26.533 & 3.330\% & 13s && RF-MoE-L & 10.617 & 1.549\% & 3s & 17.401 & 2.588\% & 14s \\
 & VaFM-POMO & \underline{16.287} & \underline{2.108\%} & 4s  & \underline{26.496} & \underline{3.187\%} & 14s  && VaFM-POMO & \underline{10.602} & \underline{1.406\%} & 4s  & \underline{17.367} & \underline{2.386\%} & 14s \\
 & VaFM-MoE-L & \textbf{16.249} & \textbf{1.868\%} & 5s  & \textbf{26.469} & \textbf{3.080\%} & 20s  && VaFM-MoE-L & \textbf{10.584} & \textbf{1.233\%} & 5s  & \textbf{17.361} & \textbf{2.350\%} & 21s \\

\midrule
\multirow{8}*{\rotatebox{90}{VRPTW}} & HGS-PyVRP & 16.032 & * & 4.6m & 25.433 & * & 9.2m & \multirow{8}*{\rotatebox{90}{OVRPTW}} & HGS-PyVRP & 10.485 & * & 4.6m & 16.900 & * & 9.2m \\
 & OR-Tools & 16.124 & 0.574\% & 4.6m & 25.923 & 1.927\% & 9.2m && OR-Tools & 10.497 & 0.114\% & 4.6m & 17.023 & 0.728\% & 9.2m \\
 & MTPOMO & 16.570 & 3.356\% & 2s & 26.403 & 3.814\% & 11s && MTPOMO & 10.851 & 3.491\% & 2s & 17.525 & 3.698\% & 11s \\
 & MVMoE & 16.455 & 2.638\% & 3s & 26.374 & 3.700\% & 14s && MVMoE & 10.760 & 2.623\% & 3s & 17.496 & 3.527\% & 15s \\
 & MVMoE-L & 16.521 & 3.050\% & 3s & 26.392 & 3.771\% & 13s && MVMoE-L & 10.797 & 2.976\% & 2s & 17.516 & 3.645\% & 14s \\
 & RF-POMO & 16.380 & 2.171\% & 2s & 26.294 & 3.385\% & 11s && RF-POMO & 10.651 & 1.583\% & 2s & 17.355 & 2.692\% & 11s \\
 & RF-MoE-L & 16.381 & 2.177\% & 3s & 26.256 & 3.236\% & 13s && RF-MoE-L & 10.656 & 1.631\% & 3s & 17.330 & 2.544\% & 14s \\
 & VaFM-POMO & \underline{16.367} & \underline{2.091\%} & 4s  & \underline{26.216} & \underline{3.077\%} & 14s  && VaFM-POMO & \underline{10.633} & \underline{1.413\%} & 4s  & \textbf{17.296} & \textbf{2.344\%} & 14s \\
 & VaFM-MoE-L & \textbf{16.335} & \textbf{1.891\%} & 5s  & \textbf{26.211} & \textbf{3.059\%} & 20s  && VaFM-MoE-L & \textbf{10.620} & \textbf{1.291\%} & 5s  & \underline{17.297} & \underline{2.347\%} & 21s \\

    \bottomrule
  \end{tabular}}
  % \end{small}
  \end{center}
\end{table*}

%% file: tables/1-sota-gain.tex
\begin{table*}[htbp]
  \caption{{Performance of VaFM and RF across tasks with different numbers of constraints. "RF" and "VaFM" correspond to RF-MoE-L and VaFM-MoE-L in Table~\ref{tab: sota}, respectively. "Number" indicates the number of constraints in each task, while "($\Delta$)" represents the improvement of VaFM over RF.}}
  \vspace{-4pt}
  \label{tab: sota-gain}
  \begin{center}
  % % \begin{small}
  \renewcommand\arraystretch{1.05}  % 0.97
  \resizebox{0.98\textwidth}{!}{ 
  \begin{tabular}{cc|cccccccccccc}
    \toprule
    % \midrule
    & & \multicolumn{6}{c}{{$N=50$}} & \multicolumn{6}{c}{$N=100$} \\
    \cmidrule(lr){3-8} \cmidrule(lr){9-14} 
     Task & Number & \multicolumn{2}{c}{{Obj. $\downarrow$}}  & \multirow{2}{*}{($\Delta$) $\uparrow$} & \multicolumn{2}{c}{{Gap $\downarrow$}} & \multirow{2}{*}{($\Delta$) $\uparrow$} & \multicolumn{2}{c}{{Obj. $\downarrow$}} & \multirow{2}{*}{($\Delta$) $\uparrow$} & \multicolumn{2}{c}{{Gap $\downarrow$}} & \multirow{2}{*}{($\Delta$) $\uparrow$} \\
     \cmidrule(lr){3-4} \cmidrule(lr){6-7} \cmidrule(lr){9-10} \cmidrule(lr){12-13} 
     & & RF & VaFM & & RF & VaFM & & RF & VaFM & & RF & VaFM & \\
     \midrule

\rowcolor[gray]{0.9}
 \multirow{1}{*}{CVRP} & \multirow{1}{*}{0}  & 10.424 & 10.421& (0.003) & 1.332\% & 1.304\% & (0.028\%) & 15.818 & 15.802 & (0.016) & 1.769\% & 1.665\% & (0.104\%) \\
% \rowcolor[gray]{0.9}
% Avg. &  &  &  & (0.003) &   && (0.028\%) &   && (0.016) &   && (0.104\%) \\

\midrule
 OVRP & \multirow{4}{*}{1} & 6.682 & 6.671 & (0.011) & 2.895\% & 2.719\% & (0.176\%) & 10.165 & 10.141 & (0.024) & 4.471\% & 4.228\% & (0.243\%) \\
 VRPB &  & 9.977& 9.948& (0.029) & 2.983\% & 2.689\% & (0.294\%) & 14.980 & 14.956 & (0.024) & 4.129\% & 3.964\% & (0.165\%) \\
 VRPL &  & 10.464 & 10.436& (0.028) & 1.317\% & 1.050\% & (0.267\%) & 15.924 & 15.921 & (0.003) & 1.835\% & 1.818\% & (0.017\%) \\
 VRPTW &  & 16.381 & 16.335& (0.046) & 2.177\% & 1.891\% & (0.286\%) & 26.256 & 26.211 & (0.045) & 3.236\% & 3.059\% & (0.177\%) \\
\rowcolor[gray]{0.9}
Avg. &  &10.876  &10.848  & (0.028) & 2.343\% 	&2.087\%   & (0.256\%) &16.831 	&16.807 & (0.024) &3.418\% 	&3.267\% & (0.151\%) \\

\midrule
 OVRPB & \multirow{6}{*}{2}  & 7.091 & 7.054& (0.037) & 2.813\% & 2.271\% & (0.542\%) & 10.773 & 10.737 & (0.036) & 4.552\% & 4.197\% & (0.355\%) \\
 OVRPL &  & 6.700 & 6.676& (0.024) & 2.919\% & 2.548\% & (0.371\%) & 10.135 & 10.134 & (0.001) & 4.388\% & 4.376\% & (0.012\%) \\
 VRPBL &  & 9.992 & 9.959& (0.033) & 3.138\% & 2.798\% & (0.340\%) & 14.982 & 14.958 & (0.024) & 4.237\% & 4.072\% & (0.165\%) \\
 OVRPTW &  & 10.656 & 10.620& (0.036) & 1.631\% & 1.291\% & (0.340\%) & 17.330 & 17.297 & (0.033) & 2.544\% & 2.347\% & (0.197\%) \\
 VRPBTW &  & 18.585 & 18.536& (0.049) & 2.301\% & 2.031\% & (0.270\%) & 30.157 & 30.090 & (0.067) & 3.990\% & 3.758\% & (0.232\%) \\
 VRPLTW &  & 16.298 & 16.249& (0.049) & 2.175\% & 1.868\% & (0.307\%) & 26.533 & 26.469 & (0.064) & 3.330\% & 3.080\% & (0.250\%) \\
\rowcolor[gray]{0.9}
Avg. &  &11.554 	&11.516   & (0.038) &2.496\% 	&2.134\% & (0.362\%) &18.318 	&18.281 & (0.038) &3.840\% 	&3.638\% & (0.202\%) \\

\midrule
 OVRPBL &  \multirow{4}{*}{3} & 7.099 & 7.062& (0.037) & 2.824\% & 2.291\% & (0.533\%) & 10.781 & 10.737 & (0.044) & 4.568\% & 4.141\% & (0.427\%) \\
 OVRPBTW &  & 11.737 & 11.703 & (0.034) & 1.268\% & 0.974\% & (0.294\%) & 19.558 & 19.511 & (0.047) & 2.040\% & 1.795\% & (0.245\%) \\
 OVRPLTW &  & 10.617 & 10.584& (0.033) & 1.549\% & 1.233\% & (0.316\%) & 17.401 & 17.361 & (0.040) & 2.588\% & 2.350\% & (0.238\%) \\
 VRPBLTW &  & 18.622 & 18.572 & (0.050) & 1.421\% & 1.150\% & (0.271\%) & 30.060 & 29.985 & (0.075) & 3.562\% & 3.305\% & (0.257\%) \\
\rowcolor[gray]{0.9}
Avg. &  &12.019 	&11.980  & (0.039) &1.766\% 	&1.412\% & (0.354\%) &19.450 	&19.399 & (0.051) &3.190\% 	&2.898\% & (0.292\%) \\

\midrule
\rowcolor[gray]{0.9}
 OVRPBLTW & 4  & 11.743 & 11.712& (0.031) & 1.259\% & 0.991\% & (0.268\%) & 19.402 & 19.359 & (0.043) & 2.089\% & 1.861\% & (0.228\%) \\

% Avg. &  &  &  & (0.031) &   & & (0.268\%) &   & & (0.043) &   & & (0.228\%) \\

    \bottomrule
    \vspace{-16pt}
  \end{tabular}}
  % \end{small}
  \end{center}
\end{table*}

%% file: tables/3-Abl-img.tex
\begin{table}[tp]
  \caption{Ablation on multi-task VRP image.}
  \vspace{-4pt}
  \label{tab: abl-att}
  \begin{center}
  % % \begin{small}
  \renewcommand\arraystretch{1.05}  % 0.97
  \setlength{\tabcolsep}{12pt}
  \resizebox{0.98\columnwidth}{!}{ 
  \begin{tabular}{cccc |cc}
    \toprule
O & B & L & TW & Obj. $\downarrow$ & Gap $\downarrow$ \\
% \midrule
% \multicolumn{4}{c|}{Baseline} & 11.429 & 2.021\% \\
\midrule
\ding{51} & \ding{51} & \ding{51} & \ding{51} & \textbf{11.409} & \textbf{1.819\%} \\
% \cmidrule{5-13}
\midrule
\ding{55} & \ding{51} & \ding{51} & \ding{51} & 11.425 & 1.861\% \\
\ding{51} & \ding{55} & \ding{51} & \ding{51} & 11.413 & 1.994\% \\
\ding{51} & \ding{51} & \ding{55} & \ding{51} & \textbf{11.409} & 1.827\% \\
\ding{51} & \ding{51} & \ding{51} & \ding{55} & 11.416 & 1.898\% \\

\midrule
\ding{55} & \ding{55} & \ding{55} & \ding{55} & 11.438 & 2.174\% \\

    % \midrule
    \bottomrule
  \end{tabular}}
  % \end{small}
  \end{center}
\vspace{-14pt}
\end{table}

%% file: tables/3-Abl-fusion.tex
\begin{table}[tp]
\color{black}
  \caption{Ablation on hybrid cross-attention fusion module.}
  \vspace{-14pt}
  \label{tab: abl-key}
  \begin{center}
  \renewcommand\arraystretch{1.05}  % 0.97
  \resizebox{1\columnwidth}{!}{ 
  \setlength{\tabcolsep}{14pt}
  \begin{tabular}{ccc | cc}
    \toprule
% \midrule
CA & LLF & HF & Obj. $\downarrow$ & \multicolumn{1}{c}{Gap $\downarrow$} \\
\midrule
\ding{51} & \ding{51} & \ding{51} & \textbf{11.409} & \textbf{1.819\%} \\
\ding{55} & \ding{51} & \ding{51} & 11.413 & 1.876\% \\
\ding{51} & \ding{55} & \ding{51} & 11.414 & 1.875\% \\
\ding{51} & \ding{51} & \ding{55} & 11.415 & 1.887\% \\
% \ding{55} & \ding{55} & 11.415 & 1.889\% \\
% \midrule
    % \midrule
    \bottomrule
  \end{tabular}}
  % \end{small}
  \end{center}
\vspace{-14pt}
\end{table}

%% file: tables/4-Abl-scale.tex
\begin{table}[t]
\color{black}
  \caption{Performance of patch embeddings from different layers. "Baseline $\times n$" denotes RF with $n$ ensemble decoders.}
  \vspace{-4pt}
  \label{tab: abl-scale}
  \begin{center}
  % % \begin{small}
  \renewcommand\arraystretch{0.97}  % 0.97
  \resizebox{0.98\columnwidth}{!}{ 
  \begin{tabular}{cccc | cc} %{cccc c cccccccc c c}
    \toprule
Layer 1 & Layer 2 & Layer 3 & Layer 4 & Obj. $\downarrow$ & Gap $\downarrow$ \\
\midrule
\multicolumn{4}{c|}{Baseline $\times$1} & 11.446 & 2.176\% \\
\ding{51} & \ding{55} & \ding{55} & \ding{55} & 11.489 & 2.620\% \\
\ding{55} & \ding{51} & \ding{55} & \ding{55} & 11.485 & 2.606\% \\
\ding{55} & \ding{55} & \ding{51} & \ding{55} & 11.437 & 2.081\% \\
\ding{55} & \ding{55} & \ding{55} & \ding{51} & 11.422 & 1.960\% \\

\midrule
\multicolumn{4}{c|}{Baseline $\times$2} & 11.429 & 2.021\% \\
\ding{51} & \ding{51} & \ding{55} & \ding{55} & 11.425 	& 1.981\% \\
\ding{51} & \ding{55} & \ding{51} & \ding{55} & 11.420 	& 1.927\% \\
\ding{51} & \ding{55} & \ding{55} & \ding{51} & 11.414 	& 1.873\% \\
\ding{55} & \ding{51} & \ding{51} & \ding{55} & 11.427 	& 1.987\% \\
\ding{55} & \ding{51} & \ding{55} & \ding{51} & 11.426 	& 1.968\% \\
\ding{55} & \ding{55} & \ding{51} & \ding{51} & 11.409  & 1.819\% \\

\midrule
\multicolumn{4}{c|}{Baseline $\times$3} & 11.432 & 2.020\% \\
\ding{51} & \ding{51} & \ding{51} & \ding{55} &11.413 	&1.858\% \\
\ding{51} & \ding{51} & \ding{55} & \ding{51} &11.411 	&1.842\% \\
\ding{51} & \ding{55} & \ding{51} & \ding{51} &11.408 	&1.809\% \\
\ding{55} & \ding{51} & \ding{51} & \ding{51} & 11.422 & 1.931\% \\

\midrule
\multicolumn{4}{c|}{Baseline $\times$4} & 11.431 & 2.015\% \\
\ding{51} & \ding{51} & \ding{51} & \ding{51} & 11.429 & 1.981\% \\

    % \midrule
    \bottomrule
  \end{tabular}}
  % \end{small}
  \end{center}
\vspace{-14pt}
\end{table}

%% file: tables/5-Dis-Gen-200.tex
\begin{table*}[tp]
  \caption{Analysis of generalization ability. "Baseline$\times$1" and "Baseline$\times$2" refer to the baseline methods with one and two decoders (the 1st and 6th rows in Table~\ref{tab: abl-scale}), respectively.
  \color{black}{VaFM\textsuperscript{\dag} is a one-decoder version, corresponding to the 5th row in Table~\ref{tab: abl-scale}.}
  \color{black}{"($\Delta$)" represents the improvement of VaFM over baselines. All methods are trained on 16 VRP variants with 50 customer nodes.}}
  \vspace{-4pt}
  \label{tab: abl-dis}
  \begin{center}
  % % \begin{small}
  \renewcommand\arraystretch{1.05}  % 0.97
  \resizebox{0.98\textwidth}{!}{ 
  \begin{tabular}{l|rrrrrrrrrrrr}
    \toprule
    % \midrule
    % & & \multicolumn{6}{c}{{$N=50$}} & \multicolumn{6}{c}{$N=100$} \\
    % \cmidrule(lr){2-7} \cmidrule(lr){8-13} 
     \multirow{2}{*}{Task} & \multicolumn{2}{c}{{Obj. $\downarrow$}}  & \multirow{2}{*}{($\Delta$) $\uparrow$} & \multicolumn{2}{c}{{Gap $\downarrow$}} & \multirow{2}{*}{($\Delta$) $\uparrow$} & \multicolumn{2}{c}{{Obj. $\downarrow$}} & \multirow{2}{*}{($\Delta$) $\uparrow$} & \multicolumn{2}{c}{{Gap $\downarrow$}} & \multirow{2}{*}{($\Delta$) $\uparrow$} \\
     \cmidrule(lr){2-3} \cmidrule(lr){5-6} \cmidrule(lr){8-9} \cmidrule(lr){11-12} 
     & Baseline$\times$1 & VaFM\textsuperscript{\dag} & & Baseline$\times$1 & VaFM\textsuperscript{\dag} & & Baseline$\times$2 & VaFM & & Baseline$\times$2 & VaFM & \\
     \midrule
\multicolumn{13}{c}{\textbf{Generalization on $N$=200}} \\
\midrule
CVRP & 24.024 & 23.582& (0.442) & 9.153\%  & 7.128\% & (2.025\%) & 23.662 & 23.842& (-0.180) & 7.501\%  & 8.309\% & (-0.808\%) \\
VRPB & 23.905 & 22.462& (1.443) & 14.010\%  & 7.121\% & (6.890\%) & 23.191 & 22.542& (0.649) & 10.606\%  & 7.478\% & (3.128\%) \\
VRPBL & 24.063 & 22.556& (1.508) & 14.290\%  & 7.124\% & (7.165\%) & 23.459 & 22.672& (0.787) & 11.419\%  & 7.663\% & (3.757\%) \\
VRPL & 24.035 & 23.576& (0.459) & 6.135\%  & 4.100\% & (2.036\%) & 23.780 & 23.878& (-0.097) & 5.000\%  & 5.434\% & (-0.434\%) \\
OVRP & 16.681 & 15.610& (1.071) & 19.815\%  & 12.067\% & (7.748\%) & 16.271 & 15.564& (0.707) & 16.788\%  & 11.723\% & (5.065\%) \\
OVRPB & 18.215 & 17.245& (0.971) & 20.801\%  & 14.324\% & (6.476\%) & 17.790 & 17.289& (0.501) & 17.864\%  & 14.623\% & (3.241\%) \\
OVRPBL & 18.257 & 17.274& (0.983) & 20.901\%  & 14.355\% & (6.545\%) & 18.307 & 17.324& (0.982) & 21.081\%  & 14.690\% & (6.391\%) \\
OVRPL & 16.721 & 15.578& (1.142) & 20.195\%  & 11.910\% & (8.285\%) & 16.852 & 15.552& (1.301) & 21.012\%  & 11.710\% & (9.302\%) \\
VRPBLTW & 50.217 & 50.150& (0.067) & 1.852\%  & 1.709\% & (0.143\%) & 50.159 & 50.342& (-0.183) & 1.730\%  & 2.106\% & (-0.375\%) \\
VRPBTW & 50.807 & 50.727& (0.080) & 1.316\%  & 1.150\% & (0.166\%) & 50.726 & 50.935& (-0.210) & 1.145\%  & 1.567\% & (-0.422\%) \\
VRPLTW & 44.121 & 44.100& (0.021) & 3.121\%  & 3.065\% & (0.056\%) & 44.130 & 44.310& (-0.180) & 3.148\%  & 3.578\% & (-0.430\%) \\
VRPTW & 43.705 & 43.726& (-0.021) & 3.234\%  & 3.285\% & (-0.051\%) & 43.694 & 43.966& (-0.272) & 3.209\%  & 3.874\% & (-0.665\%) \\
OVRPBLTW & 33.454 & 33.503& (-0.049) & 4.295\%  & 4.446\% & (-0.151\%) & 33.318 & 33.810& (-0.492) & 3.877\%  & 5.411\% & (-1.534\%) \\
OVRPBTW & 33.454 & 33.492& (-0.038) & 4.379\%  & 4.489\% & (-0.110\%) & 33.318 & 33.854& (-0.535) & 3.960\%  & 5.617\% & (-1.657\%) \\
OVRPLTW & 29.927 & 30.061& (-0.134) & 8.661\%  & 9.134\% & (-0.473\%) & 29.863 & 30.323& (-0.460) & 8.438\%  & 10.096\% & (-1.658\%) \\
OVRPTW & 29.897 & 29.993& (-0.096) & 8.751\%  & 9.092\% & (-0.340\%) & 29.818 & 30.328& (-0.511) & 8.475\%  & 10.312\% & (-1.837\%) \\
% \midrule
\rowcolor[gray]{0.9}
Avg. & 30.093 & 29.602& (0.491) & 10.057\%  & 7.156\% & (2.901\%) & 29.896 & 29.783& (0.113) & 9.078\%  & 7.762\% & (1.316\%) \\
\midrule
\multicolumn{13}{c}{\textbf{Generalization on 8 unseen vraints}} \\
\midrule
OVRPMB & 6.981 & 6.972& (0.009) & 14.381\%  & 14.252\% & (0.130\%) & 6.945 & 6.947& (-0.001) & 13.802\%  & 13.822\% & (-0.019\%) \\
OVRPMBL & 6.991 & 6.975& (0.016) & 14.423\%  & 14.159\% & (0.264\%) & 6.941 & 6.950& (-0.010) & 13.590\%  & 13.748\% & (-0.157\%) \\
VRPMB & 9.910 & 9.864& (0.046) & 9.430\%  & 8.908\% & (0.522\%) & 9.848 & 9.860& (-0.011) & 8.730\%  & 8.854\% & (-0.124\%) \\
VRPMBL & 9.929 & 9.899& (0.031) & 9.112\%  & 8.761\% & (0.351\%) & 9.879 & 9.885& (-0.006) & 8.541\%  & 8.610\% & (-0.068\%) \\
OVRPMBLTW & 11.082 & 11.081& (0.001) & 6.291\%  & 6.282\% & (0.009\%) & 11.046 & 11.050& (-0.004) & 5.950\%  & 5.967\% & (-0.017\%) \\
OVRPMBTW & 11.100 & 11.109& (-0.009) & 6.112\%  & 6.201\% & (-0.088\%) & 11.083 & 11.083& (0.001) & 5.937\%  & 5.942\% & (-0.005\%) \\
VRPMBLTW & 17.241 & 17.226& (0.015) & 7.923\%  & 7.825\% & (0.097\%) & 17.193 & 17.187& (0.006) & 7.597\%  & 7.564\% & (0.033\%) \\
VRPMBTW & 17.282 & 17.256& (0.026) & 8.004\%  & 7.863\% & (0.141\%) & 17.245 & 17.222& (0.023) & 7.760\%  & 7.623\% & (0.137\%) \\
% \midrule
\rowcolor[gray]{0.9}
Avg. & 11.314 & 11.298& (0.017) & 9.460\%  & 9.281\% & (0.178\%) & 11.273 & 11.273& (0.000) & 8.301\%  & 9.016\% & (-0.715\%) \\
\midrule
\multicolumn{13}{c}{\textbf{Generalization on real-world CVRPLib dataset}} \\
\midrule
Group A & 106222 & 107963& (-1741) & 1.845\%  & 3.373\% & (-1.528\%) & 106856 & 108033& (-1178) & 2.466\%  & 3.416\% & (-0.950\%) \\
Group B & 99026 & 99443& (-417) & 2.669\%  & 3.025\% & (-0.356\%) & 99126 & 99552& (-426) & 2.871\%  & 3.169\% & (-0.298\%) \\
Group E & 74864 & 75145& (-282) & 2.719\%  & 2.832\% & (-0.113\%) & 74809 & 75209& (-400) & 2.648\%  & 2.849\% & (-0.201\%) \\
Group F & 79300 & 79167& (133) & 10.868\%  & 10.763\% & (0.105\%) & 78833 & 78267& (567) & 12.231\%  & 10.091\% & (2.140\%) \\
Group M & 119640 & 117360& (2280) & 10.104\%  & 8.078\% & (2.026\%) & 119240 & 119680& (-440) & 9.830\%  & 10.222\% & (-0.392\%) \\
Group P & 60313 & 60675& (-363) & 2.810\%  & 3.342\% & (-0.532\%) & 60571 & 60613& (-42) & 3.919\%  & 3.157\% & (0.762\%) \\
Group X<251 & 3442436 & 3433130& (9306) & 8.415\%  & 8.222\% & (0.193\%) & 3435621 & 3462430& (-26809) & 7.871\%  & 9.301\% & (-1.430\%) \\
Group 251<X<501 & 6517137 & 6350520& (166617) & 14.980\%  & 12.474\% & (2.506\%) & 6424537 & 6674926& (-250389) & 12.983\%  & 19.058\% & (-6.075\%) \\
\rowcolor[gray]{0.9}
Avg. & 1312367 & 1290425& (21942) & 6.801\%  & 6.514\% & (0.288\%) & 1299949 & 1334839& (-34890) & 6.852\%  & 7.658\% & (-0.805\%) \\

    \bottomrule
  \end{tabular}}
  % \end{small}
  \end{center}
  \vspace{-8pt}
\end{table*}

%% file: 5-dis.tex
\section{Discussion}
\subsection{Generalization Ability}
To investigate generalization capacity, we employ two variants of VaFM and evaluate their performance on large-scale VRPs, unseen variants, and a real-world CVRPLib dataset. VaFM\textsuperscript{\dag} is a single-decoder version of VaFM, corresponding to the 5th row in Table~\ref{tab: abl-scale}. Table~\ref{tab: abl-dis} presents the results of the two variants and their corresponding baselines. All models are trained on 16 VRP variants with $N$=50 and are subsequently tested on these settings in a zero-shot manner.

\subsubsection{Large-scale Data} 
The models are tested on 16 variants with 200 customer nodes, and the upper block of Table~\ref{tab: abl-dis} presents the scores for these variants along with the averaged result. These results show that VaFM\textsuperscript{\dag} consistently outperforms Baseline$\times$1 on variants without TW.
In terms of averaged performance, both versions surpass their corresponding baselines, with VaFM\textsuperscript{\dag} achieving a 2.901\% gap reduction, while VaFM achieves a smaller 1.316\% reduction.
\subsubsection{Unseen Constraints}
We follow~\cite{berto2024routefinder} and adopt mixed backhaul (MB), which relaxes visit order restrictions, allowing linehaul and backhaul customers to be mixed along a route in any configuration. This setting results in 8 variants, with the corresponding scores and their average values presented in the middle block of Table~\ref{tab: abl-dis}. These results show a similar observation to the large-scale generalization. VaFM\textsuperscript{\dag} consistently outperforms Baseline$\times$1 on variants without TW. VaFM\textsuperscript{\dag} achieves a 0.178\% reduction in the average gap, while the two-decoder version shows a 0.715\% increase.
\subsubsection{Real-world Dataset}
We evaluate 338 instances with diverse node distributions and demands. These instances are grouped into \{A, B, E, F, M, P, X\}, with a maximum scale of 500 customer nodes, and their average scores are reported at the bottom of Table~\ref{tab: abl-dis}. These results indicate that both versions of VaFM reduce the Gap in some groups but exhibit instability in others. The average scores show that the gap of the one-decoder version is reduced by 0.288\%, while the two-decoder version results in a 0.805\% increase.

\subsubsection{Limitations and Future Work}
\color{black}
As shown in Table~\ref{tab: abl-dis}, we observe that both VaFM and VaFM\textsuperscript{\dag} exhibit a slightly larger gap on the $N$=200 TW tasks. We attribute it to the moderate discriminative capacity of a single color. When the training size $N$=50 scales to $N$=200 during testing, the TW differences among nodes become less significant, resulting in minimal color contrast. In future work, we plan to revise the image by using two colors (\eg, yellow for the largest TW and blue for the smallest) to mitigate this limitation.
In addition, VaFM shows an increased average Gap on unseen constraints and the CVRPLib dataset compared to its baseline. While VaFM\textsuperscript{\dag} can mitigate this issue, it still does not achieve consistent improvement across all individual tasks. In future work, we plan to integrate textual instructions into the images and develop a multi-modal large language model to enhance generalization performance.
\color{black}
\input{tables/5-Dis-Comp}

\subsection{Analysis of Model Complexity}

Table~\ref{tab: abl-dis-complexity} presents the model complexity of both the baseline and our methods. "Time" refers to the cumulative training time, while "\# Params" indicates the number of trainable model parameters. Regarding inference cost metrics, we consider the maximum memory usage and latency. All models are trained and evaluated on 16 VRP variants with $N$=50 using an RTX 3090 GPU. These metrics are evaluated with a batch size of 150, and "Latency" refers to the average time across the 16 variants, each consisting of 1000 instances. 
\color{black}
The results show that VaFM requires a higher model cost to achieve a smaller gap. To ensure a fair comparison, we scale up the baseline by using a heavier encoder (HE), which increases the encoder’s feature dimension from 128 to 512, denoted as "Baseline$\times$2 + HE". Results in Table~\ref{tab: abl-dis-complexity} show that adding a heavy encoder even enlarges the gap (2.021\% $\xrightarrow{}$ 2.281\%). In contrast, under the comparable training time, parameters, and latency with "Baseline$\times$2 + HE", VaFM achieves the smallest Gap of 1.819\%. This confirms that the performance gain stems from the vision branch rather than model scaling. To further reduce computational cost, we incorporate FlashAttention~\cite{dao2022flashattention} and optimize the matrix operations in Eq.~\ref{eq: patch_emb}, forming "VaFM + SpeedUp". As shown in Table~\ref{tab: abl-dis-complexity}, this variant achieves a 0.3s speedup and saves 11.9 GB of memory, which motivates us to develop more efficient implementations in the future.
\color{black}

\color{black}
\input{tables/5-dis-TE}
\subsection{Comparison with the New Version of RouteFinder}

RouteFinder~\cite{berto2024routefinder} has released a new version~\cite{berto2025routefinder}, which identifies RF-TE as the best-performing variant and extends it to the unseen multi-depot (MD) constraint in a zero-shot manner. This version evaluates newly generated instances with a different random seed, so a direct comparison between RF-TE and the results in Table~\ref{tab: sota} would be unfair. To address this, we integrate VaFM into the new variant and evaluate its performance on the newly generated test data. Table~\ref{tab: dis-te} reports the gap and inference time across all tasks, both with and without the MD constraint. We observe that VaFM-TE outperforms RF-TE in 11 out of 16 tasks without MD, and consistently surpasses it across tasks with MD. This indicates that VaFM is a plug-and-play method that can be integrated into various graph-based VRP solvers, while achieving strong performance and good generalization ability.
\color{black}

%% file: tables/5-Dis-Comp.tex
% \begin{table}[tp]
%   \caption{Analysis of model complexity.}
%   \vspace{-6pt}
%   \label{tab: abl-dis-complexity}
%   \begin{center}
%   % % \begin{small}
%   \renewcommand\arraystretch{1.05}  % 0.97
%   \resizebox{0.98\columnwidth}{!}{ 
%   \begin{tabular}{c| rr|rr} %{cccc c cccccccc c c}
%     \toprule
% Metrics & Baseline$\times$1 & VaFM$\times$1 & Baseline$\times$2 & VaFM$\times$2 \\
% \midrule
% Gap & 2.176\% & 1.960\% & 2.021\% & 1.819\% \\
% \midrule
% Training Time (h) & 25.283 & 17.083 & 31.533 & 22.317 \\
% Memory (GB) & 2.459 & 7.844 & 2.498 & 19.665 \\
% \# Params (M) & 3.730 & 15.747 & 3.900 & 16.110 \\
% Latency (s) & 2.510 & 3.712 & 3.812 & 5.274 \\

%     % \midrule
%     \bottomrule
%   \end{tabular}}
%   % \end{small}
%   \end{center}

% \end{table}

\begin{table}[tp]
\color{black}
    \caption{Analysis of model complexity.}
  % \vspace{-6pt}
  \label{tab: abl-dis-complexity}
    \centering
    \setlength{\tabcolsep}{10pt}
    \resizebox{1\columnwidth}{!}{%
    \begin{tabular}{@{}l cccc c@{}}
        \toprule
        \multirow{2}{*}{Method} & \makecell[c]{Time $\downarrow$} & \makecell[c]{Memory $\downarrow$} & \makecell[c]{\# Params $\downarrow$} & \makecell[c]{Latency $\downarrow$} & \multirow{2}{*} {Gap $\downarrow$}\\
        & \makecell[c]{(h)} & \makecell[c]{(GB)} & \makecell[c]{(M)} & \makecell[c]{(s)}  \\
        \midrule
        Baseline$\times$1 & 25.6 & 2.5 & 3.7 & 2.5 & 2.176\% \\
        Baseline$\times$2 & 32.6 & 2.5 & 3.9 & 3.8 & 2.021\% \\
        Baseline$\times$2 + HE & 58.4 & 2.6 & 19.4 & 4.4 & 2.281\% \\
        % Baseline$\times$2 + HEHD & 0.0 & 3.1 & 24.7 & 7.5 & 0.000\% \\
        \midrule
        VaFM\textsuperscript{\dag} & 30.3 & 7.8 & 15.7 & 3.7 & 1.960\% \\
        VaFM & 44.6 & 19.7 & 16.1 & 5.1 & 1.819\% \\
        VaFM + SpeedUp & 44.6 & 7.8 & 16.1 & 4.8 & 1.817\% \\
        \bottomrule
    \end{tabular}
    }
    \vspace{-0.2cm}
\end{table}

%% file: tables/5-dis-TE.tex
\begin{table}[tp]
\color{black}
  \caption{Comparison with the New Version of RouteFinder. The best results are highlighted with gray shading.}
  \vspace{-14pt}
  \label{tab: dis-te}
  \begin{center}
  \setlength{\tabcolsep}{6pt}
  \renewcommand\arraystretch{1.05}  % 0.97
  \resizebox{1\columnwidth}{!}{ 
  \begin{tabular}{l|cccc | cccc}
    \toprule
    & \multicolumn{4}{c|}{w/o MD}  & \multicolumn{4}{c}{w/ MD} \\
    \cmidrule(lr){2-5} \cmidrule(lr){6-9} 
     Task& \multicolumn{2}{c}{{RF-TE}}  & \multicolumn{2}{c|}{VaFM-TE} & \multicolumn{2}{c}{{RF-TE}}  & \multicolumn{2}{c}{VaFM-TE} \\
     \cmidrule(lr){2-3} \cmidrule(lr){4-5} \cmidrule(lr){6-7} \cmidrule(lr){8-9}
     & Gap $\downarrow$ & Time $\downarrow$ & Gap $\downarrow$ & Time $\downarrow$ & Gap $\downarrow$ & Time $\downarrow$ & Gap $\downarrow$ & Time $\downarrow$ \\
     \midrule
CVRP & \cellcolor{gray!20} 1.274\% & 2.1s & 1.285\% & 4.1s & 56.968\% & 3.2s & \cellcolor{gray!20} 35.873\% & 4.5s \\
OVRP & \cellcolor{gray!20} 2.687\% & 1.7s & 2.735\% & 3.7s & 29.061\% & 2.1s & \cellcolor{gray!20} 23.189\% & 3.8s \\
VRPB & 2.989\% & 1.7s & \cellcolor{gray!20} 2.971\% & 3.6s & 60.980\% & 2.3s & \cellcolor{gray!20} 48.303\% & 4.1s \\
VRPL & 1.502\% & 1.7s & \cellcolor{gray!20} 1.489\% & 3.7s & 57.386\% & 2.4s & \cellcolor{gray!20} 42.997\% & 4.7s \\
OVRPB & 2.479\% & 1.8s & \cellcolor{gray!20} 2.355\% & 3.8s & 31.870\% & 2.2s & \cellcolor{gray!20} 22.889\% & 4.0s \\
OVRPL & 2.721\% & 1.7s & \cellcolor{gray!20} 2.693\% & 3.7s & 28.944\% & 2.1s & \cellcolor{gray!20} 22.910\% & 3.8s \\
VRPBL & \cellcolor{gray!20} 3.803\% & 1.8s & 3.898\% & 3.7s & 61.198\% & 2.5s & \cellcolor{gray!20} 55.733\% & 4.8s \\
VRPTW & 2.077\% & 1.9s & \cellcolor{gray!20} 2.047\% & 3.8s & 48.943\% & 2.5s & \cellcolor{gray!20} 33.720\% & 4.5s \\
OVRPBL & 2.508\% & 1.8s & \cellcolor{gray!20} 2.420\% & 3.8s & 31.759\% & 2.2s & \cellcolor{gray!20} 23.032\% & 3.9s \\
OVRPTW & \cellcolor{gray!20} 1.326\% & 2.0s & 1.352\% & 4.0s & 34.903\% & 2.3s & \cellcolor{gray!20} 28.642\% & 4.3s \\
VRPBTW & 1.676\% & 1.9s & \cellcolor{gray!20} 1.606\% & 3.9s & 37.347\% & 2.5s & \cellcolor{gray!20} 33.002\% & 4.7s \\
VRPLTW & 2.454\% & 1.9s & \cellcolor{gray!20} 2.407\% & 3.9s & 51.254\% & 2.6s & \cellcolor{gray!20} 35.331\% & 4.5s \\
OVRPBTW & 1.151\% & 2.0s & \cellcolor{gray!20} 1.092\% & 4.1s & 32.325\% & 2.4s & \cellcolor{gray!20} 28.589\% & 4.5s \\
OVRPLTW & \cellcolor{gray!20} 1.340\% & 2.0s & 1.351\% & 4.0s & 34.716\% & 2.3s & \cellcolor{gray!20} 28.498\% & 4.3s \\
VRPBLTW & 1.877\% & 1.9s & \cellcolor{gray!20} 1.829\% & 4.0s & 40.597\% & 2.7s & \cellcolor{gray!20} 34.582\% & 4.9s \\
OVRPBLTW & 1.151\% & 2.0s & \cellcolor{gray!20} 1.102\% & 4.1s & 32.202\% & 2.4s & \cellcolor{gray!20} 28.235\% & 4.5s \\

    \bottomrule
  \end{tabular}}
  \end{center}
  \vspace{-8pt}
\end{table}

%% file: 6-con.tex
\section{Conclusion}
In this work, we propose a vision-assisted foundation model (VaFM) to address the challenges of solving multi-task VRPs with complex constraints. 
By integrating vision modality with graph-based models, VaFM captures patch-level semantics from the multi-task VRP image, providing spatial demand distributions or
service time patterns to enhance node embeddings. 
Both the hybrid cross-attention fusion module and the constraint-aware auxiliary task contribute to enhancing the performance of VaFM. The hybrid cross-attention fusion module enables the model to adaptively adjust its receptive fields for different tasks, while the constraint-aware auxiliary task utilizes binary cross-entropy loss to balance constraint learning.
Our experimental results demonstrate the superior performance of VaFM compared to existing SOTA methods, particularly for VRP variants with complex constraints. 
We further analyze the generalization ability and model complexity, identifying VaFM's limitations in handling unseen constraints and real-world datasets. To address this issue, we provided the alternative one-decoder version of VaFM with better generalization. However, we plan to explore effective methods to exclusively enhance the original VaFM’s generalization in the future.

% \colorr{(Better have some future works.)}